\documentclass[runningheads]{llncs}

\usepackage[mobile]{eccv}

\usepackage{eccvabbrv}

\usepackage{graphicx}
\usepackage{booktabs}

\usepackage[accsupp]{axessibility}  % Improves PDF readability for those with disabilities.

\usepackage[pagebackref,breaklinks,colorlinks,citecolor=eccvblue]{hyperref}
\usepackage{orcidlink}
\usepackage{multicol}
\usepackage{multirow}

\begin{document}

\title{DMAD: Dual Memory Bank for Real-World Anomaly Detection} 


\author{Jianlong Hu\inst{1,2} \and
Xu Chen\inst{2} \and
Zhenye Gan\inst{2} \and
Jinlong Peng\inst{2} \and
Shengchuan Zhang\inst{1*} \and
Jiangning Zhang\inst{2} \and
Yabiao Wang\inst{2} \and
Chengjie Wang\inst{2} \and
Liujuan Cao\inst{1} \and
Rongrong Ji\inst{1}
}

\authorrunning{F.~Author et al.}

\institute{
Key Laboratory of Multimedia Trusted Perception and Efficient Computing, Ministry of Education of China,
School of Informatics, Xiamen University
 \and
Youtu Lab, Tencent
}

\maketitle

\begin{abstract}
    Training a unified model is considered to be more suitable for practical industrial anomaly detection scenarios due to its generalization ability and storage efficiency. However, this multi-class setting, which exclusively uses normal data, overlooks the few but important accessible annotated anomalies in the real world. To address the challenge of real-world anomaly detection, we propose a new framework named Dual Memory bank enhanced representation learning for Anomaly Detection (DMAD). This framework handles both unsupervised and semi-supervised scenarios in a unified (multi-class) setting. DMAD employs a dual memory bank to calculate feature distance and feature attention between normal and abnormal patterns, thereby encapsulating knowledge about normal and abnormal instances. This knowledge is then used to construct an enhanced representation for anomaly score learning. We evaluated DMAD on the MVTec-AD and VisA datasets. The results show that DMAD surpasses current state-of-the-art methods, highlighting DMAD's capability in handling the complexities of real-world anomaly detection scenarios. The code will be made available.
\end{abstract}
\section{Introduction}
\begin{figure}
    \centering
    \includegraphics[width=1.0\textwidth]{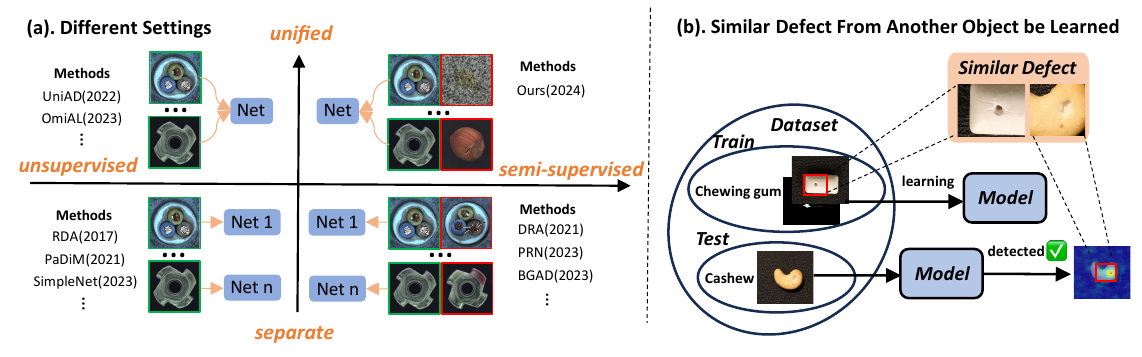}
    \caption{\textbf{(a). Comparison of Anomaly Detection's Settings.} The current research trend is moving from one-class to unified (multi-class) and from unsupervised to semi-supervised, which is more practical. We propose a unified with a few annotated anomalies setting, filling the gap in this field. \textbf{(b).} Defects in various objects often exhibit visual similarities. Therefore, defect data from one object can aid the model in detecting similar defects in other objects within a \textbf{unified semi-supervised setting}.}
  \label{fig:intro}
  \vspace{-0.5cm}
\end{figure}

Image anomaly detection, also known as surface defect detection, is a process that classifies images into normal and abnormal categories, pinpointing anomalies at the pixel level. Deep learning-based anomaly detection methods have demonstrated superior efficiency compared to manual anomaly localization.
Prevailing approaches typically train a unique model for each object, as illustrated in the bottom-left of \Cref{fig:intro} (a). However, this approach results in increased storage consumption as the number of object categories grows rapidly. To address this issue, UniAD~\cite{you2022unified} proposed a multi-class setting that leverages all normal data from objects to train a unified model, as depicted in the top-left of \Cref{fig:intro} (a).
Furthermore, under the assumption that anomaly data is unavailable, current anomaly detection methods~\cite{zhou2017anomaly, defard2021padim, liu2023simplenet, wang2023multimodal} primarily depend on unsupervised learning. They explicitly define the boundaries of normal data, as shown on the left of \Cref{fig:intro} (a). However, these boundaries may lack sufficient accuracy due to the absence of real anomaly data during training, especially when dealing with minor defects.
Recent studies~\cite{pang2021explainable, ding2022catching, zhang2023prototypical, yao2023explicit} have suggested the feasibility of obtaining a handful of anomalies in real-world situations. These semi-supervised approach, shown on the bottom-right of \Cref{fig:intro} (a), can aid the model in predicting potential anomaly patterns and enhancing its performance.
In our proposition, for real-world anomaly detection, training a unified model is more compatible. Furthermore, beyond unsupervised scenarios, semi-supervised scenarios also need to be considered, where a few annotated anomalies are available. We denote this new unified setting with a few annotated anomalies as a \textbf{unified semi-supervised setting}, which is illustrated in the top-right of \Cref{fig:intro} (a). This new setting is filling the research gap.

Under a unified semi-supervised setting, the model utilizes data from all object categories, including normal data, visible abnormal data, and their corresponding supervision information for training. This approach offers several benefits. Firstly, it aligns more closely with real-world scenarios, as there are always some annotated anomalies available for use. Secondly, this setting is more unified and convenient, as all data, including annotations, are utilized by a single model. Thirdly, we observe visually common defects across different object categories, as depicted in \Cref{fig:intro} (b). By employing this setting for anomaly detection, these similar defects can provide additional advantages during the training process. The challenge of this setting lies in accurately modeling a multi-class distribution while avoiding overfitting to visible anomalies and effectively utilizing the available seen anomalies. To address this, we propose a novel framework: Dual Memory bank enhanced representation learning for Anomaly Detection (DMAD). DMAD is not only suitable for the unified semi-supervised setting, but also compatible with the general unified (multi-class) setting, where no annotated anomalies are available. This makes it well-suited for real-world anomaly detection, where anomalies may not be initially accessible. However, as the system operates, certain annotated anomalies may emerge for use.

Specifically, DMAD first employs a patch feature encoder to extract the patched feature. To establish a unified decision boundary, a dual memory bank, comprising a normal and abnormal memory bank, is constructed. In the unsupervised scenario, a pseudo abnormal feature set, generated from the feature fusion of normal and outlier features, is utilized as the abnormal memory bank. In semi-supervised scenarios, to mitigate data imbalance, we employ an anomaly center sampling strategy. This strategy expands the abnormal memory bank, supplementing the observed anomalies and the pseudo abnormal feature set. The dual memory bank calculates the distance and cross-attention between patched features and their nearest features in both memory banks, creating a knowledge base for normal and abnormal data. The feature itself, along with its two calculated knowledge components, comprise an enhanced representation. Finally, we use a Multilayer Perceptron (MLP) to learn a mapping between this representation and the anomaly score.

In summary, our contributions are as follows:

1. To address real-world anomaly detection, which encompasses both a general unified setting and a unified semi-supervised setting, we propose a novel dual memory bank-based framework named DMAD. DMAD employs a normal memory bank and a variable abnormal memory bank to handle both scenarios.

2. To effectively utilize the dual memory bank, we introduce a knowledge enhancement module. This module calculates a distance and a cross-attention to form a knowledge representation, facilitating improved anomaly score learning.

3. We conduct extensive experiments on the MVTec-AD and VisA datasets. The results demonstrate that our model significantly outperforms current state-of-the-art competitors in various settings. 
% These results also establish new benchmarks for future research in this crucial emerging field.
\section{Related Work}
\label{sec:formatting}

\subsection{Unsupervised Anomaly Detection}
Most anomaly detection methods assume that only normal data can be accessed during training. To this end, various unsupervised techniques are proposed and can be roughly divided into three categories:
% including the reconstruction-based methods, memory bank-based methods, and the data augmentation-based methods. 
% \textbf{Reconstruction-based methods} are based on the assumption that models trained only on normal samples can accurately reconstruct normal regions but fail to handle anomalous regions. Preliminary attempts under different generative models are made, including Auto-Encoder~\cite{bergmann2018improving,chen2017outlier,zhou2017anomaly}, Variational Auto-Encoder~\cite{liu2020towards,dehaene2020iterative}, and Generative Adversarial Net~\cite{sabokrou2018adversarially,schlegl2017unsupervised,schlegl2019f}. 
\textbf{1) Reconstruction-based approaches} assume models trained solely on normal samples accurately reconstruct normal regions but struggle with anomalous ones. Initial attempts involve various generative models, including Auto-Encoder~\cite{bergmann2018improving,chen2017outlier,zhou2017anomaly}, Variational Auto-Encoder~\cite{liu2020towards,dehaene2020iterative,vitad}, Generative Adversarial Net~\cite{sabokrou2018adversarially,schlegl2017unsupervised,schlegl2019f,liang2023omni}, and Diffusion-based methods~\cite{diad}.
%These methods are straightforward because the anomaly map is directly from pixel-wise difference between the input image and the reconstruct one. But their performance is limited because some anomaly can be reconstructed also by the generalization of model. 
\textbf{2) Memory bank-based methods} extract and save representations of normal images from pre-trained deep neural networks and then detect anomalies by comparing the features. SPADE~\cite{cohen2020sub} introduces the multi-resolution semantic pyramid anomaly detection framework. 
%, which aims to estimate dense pixel-level correspondence between the target and the normal. uses correspondences based on a multi-resolution feature
PaDiM~\cite{defard2021padim} uses multivariate Gaussian distributions to get a
probabilistic representation of the normal class. PatchCore~\cite{roth2022towards} proposes greedy coreset subsampling to lighten the memory bank. 
% \textbf{Data augmentation-based methods} produces anomalies on anomaly-free images and converts anomaly detection to supervised learning. CutPaste~\cite{li2021cutpaste} proposes a simple data augmentation strategy that cuts an image patch and pastes at a random location of a large image. DRÆM~\cite{zavrtanik2021draem} uses Perlin noise to generate just-out-of-distribution appearances. NSA~\cite{schluter2022natural} integrates Poisson image editing to seamlessly blend scaled patches from different images. SimpleNet~\cite{liu2023simplenet} adds Gaussian noise in the feature space rather than directly on the images and achieve SOTA performance recently.
\textbf{3) Data augmentation methods} create anomalies on anomaly-free images, transforming anomaly detection into supervised learning~\cite{li2021cutpaste,anomalydiffusion}. CutPaste~\cite{li2021cutpaste} introduces a straightforward strategy, cutting an image patch and pasting it randomly onto a larger image. DRÆM~\cite{zavrtanik2021draem} employs Perlin noise to generate just-out-of-distribution appearances. NSA~\cite{schluter2022natural} integrates Poisson image editing to seamlessly blend scaled patches from different images. SimpleNet~\cite{liu2023simplenet} introduces Gaussian noise in the feature space, achieving recent state-of-the-art (SOTA) performance.
These approaches focus on the problem that no anomalous samples are accessed in training stage. But some challenging anomalies are quite difficult to distinguish due to the lack of true anomalies knowledge. 

%-------------------------------------------------------------------------
% \subsection{Semi-Supervised Anomaly Detection}
% In practice however, a small amount of anomalous data is also obtainable in addition to normal data. Under this premise, some semi-supervised anomaly detection methods have proposed to make full use of the few but important anomalous images recently. DeepSAD~\cite{ruff2019deep} firstly focuses on this semi-supervised AD setting and introduce an information-theoretic framework. DRA~\cite{ding2022catching} learns disentangled representations of defects by seen and pseudo anomalies. PRN~\cite{zhang2023prototypical} learns feature
% residuals between anomalous and normal patterns to accurately detect anomalous regions. BGAD~\cite{yao2023explicit} proposes a boundary guided semi-push-pull contrastive learning mechanism based on conditional normalizing flow.
\subsection{Semi-Supervised Anomaly Detection}
In practice, a small amount of anomalous data is obtainable alongside normal data. Some semi-supervised anomaly detection methods aim to fully utilize these scarce but crucial anomalous images. DeepSAD~\cite{ruff2019deep} initially focuses on this semi-supervised AD setting and introduces an information-theoretic framework. DRA~\cite{ding2022catching} learns disentangled representations of defects through seen and pseudo anomalies. PRN~\cite{zhang2023prototypical} learns feature residuals between anomalous and normal patterns for accurate detection. BGAD~\cite{yao2023explicit} proposes a boundary-guided semi-push-pull contrastive learning mechanism based on conditional normalizing flow.

With the help of anomalous samples, these semi-supervised methods have achieved better results compared to unsupervised methods. However, they are designed for one-class-one-model scenarios, which is cumbersome when the number of product categories increases. Additionally, they are unable to leverage knowledge between different products with similar defects.

%-------------------------------------------------------------------------
\subsection{Multi-class Unsupervised Anomaly Detection}
Most methods propose to train separate models for different classes of objects. However, in real-world application scenarios, the number of intra-class or intre-class may be quite large and separate-trained methods become uncongenial and memory-consuming. Among this, UniAD~\cite{you2022unified} firstly proposes to detect anomalies from different objects with a unified framework. Meanwhile, UniAD introduces a layer-wise query decoder to the transformer framework to weaken the “identical shortcut” problem of reconstruction-based methods especially in unified setting. OmniAL~\cite{zhao2023omnial} improves anomaly synthesis, reconstruction, and localization by using panel-guided synthetic anomaly data.
These methods are convenient in practical applications because they use a unified model that can cover all objects. However, they cannot utilize defects to enhance their performance. In this paper, we propose DMAD which can model a multi-class distribution and effectively utilize the available anomalies as well.
\begin{figure}
    \centering
    \includegraphics[width=1\textwidth]{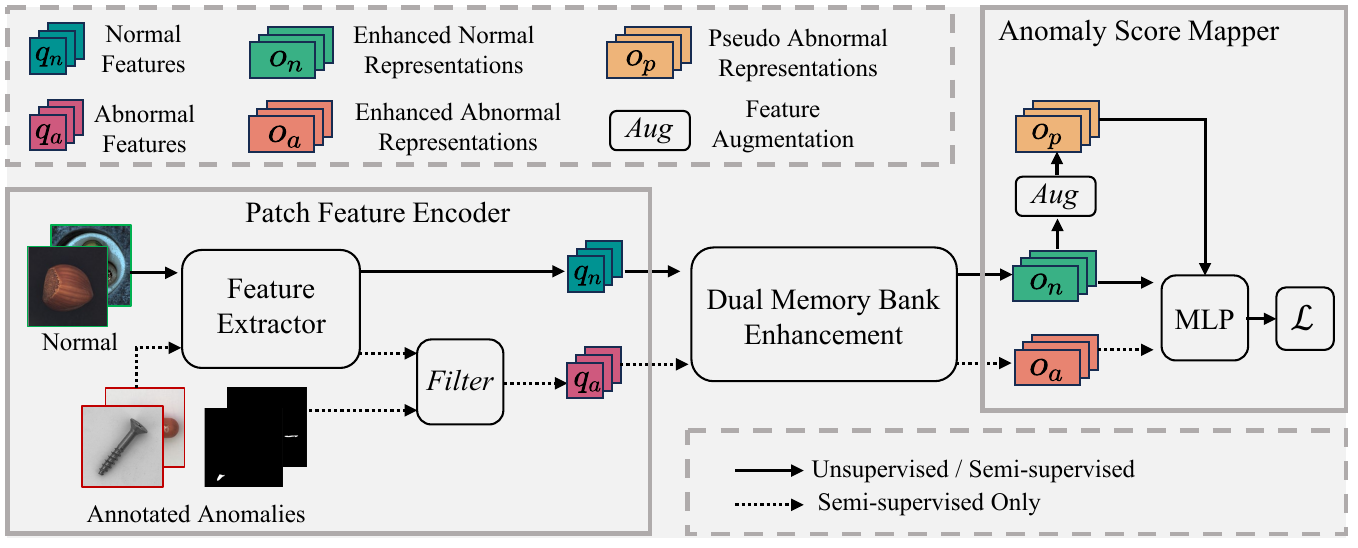}
    \caption{\textbf{Overview of DMAD}. DMAD is a unified framework that accommodates both unsupervised and semi-supervised scenarios. During training, normal images and seen annotated anomalies are input into the Feature Extractor to obtain patched features. For the anomalies, a $Filter$ is utilized to isolate the anomalous parts. Subsequently, a Dual Memory Bank Knowledge Enhancement is employed to obtain enhanced representations of both normal and abnormal. A more detailed illustration is depicted in \Cref{fig:framework_detail}.  A pseudo abnormal representation is generated by utilized a feature augmentation~\cite{liu2023simplenet}. Ultimately, these normal representations, abnormal representations, and pseudo abnormal representations are used to train a MLP. }
  \label{fig:framework}
\end{figure}

\section{Methodology}

\textbf{Problem Statement.}
In practical industrial scenarios, training a unified model is viewed as more compatible and storage-efficient. The unified anomaly detection system is confronted with two situations simultaneously: a general unified (multi-class) setting~\cite{you2022unified}and an unified setting with a few annotated anomalies, which can be referred to as unified semi-supervised setting. This dual situation arises depending on the availability of anomalies. In the general unified setting, the model's training set is denoted as $\mathcal{X}_{\text{train}}=\{\mathcal{X}_n^i\}_{i=1}^{M}$. Here, $M$ represents the number of objects in the dataset, while $\mathcal{X}_n$ signifies the normal data. When some annotated anomalies become available, the setting transitions to unified semi-supervised, and the training set changes to $\mathcal{X}_{\text{train}}=\{\mathcal{X}_n^i\}_{i=1}^{M} \cup \{\mathcal{X}_{a_s}^i\}_{i=1}^{M}$. $\mathcal{X}_{a_s}$ represents the seen annotated anomalies. The objective is to train a unified neural network, denoted as $m: \mathcal{X} \to \mathbb{R}  $, capable of assigning higher anomaly scores to anomalies than to normal instances. For real-world anomaly detection, the model needs to accommodate both settings.

\noindent\textbf{Overview.}
We propose a novel framework, named Dual Memory bank enhanced representation learning for Anormaly Detection (DMAD), designed to tackle the challenges of real-world anomaly detection. DMAD is a unified model that not only utilizes normal data for training but also effectively employs accessible anomalies. An overview of DMAD can be found in \Cref{fig:framework}. DMAD primarily consists of three components: the Patch Feature Encoder (\Cref{sec:PFE}), Dual Memory Bank-based Knowledge Enhancement (\Cref{sec:DMBKE}), and Anomaly Score Mapper (\Cref{sec:ASP}). We will subsequently provide a detailed introduction to these three components.

\subsection{Patch Feature Encoder}
The Patch Feature Encoder consists of a Feature Extractor $F_{\Phi}:x \to q$, and an optional feature filter operation $Filter$. The Feature Extractor $F_{\Phi}$ is employed to extract the patched features from an image, which includes a pre-trained backbone and an aggregation operation~\cite{roth2022towards}. The training image is represented as  $x \in \mathbb{R}^{3 \times H \times W}$, and the patched feature denoted as $q \in \mathbb{R}^{N \times C}$. Here, $N=H_0 \times W_0$ represents the number of patches, $H_0$ and $W_0$ denote the height and width of the feature respectively, and $C$ signifies the feature’s channel. 

For a general unified setting, where there are only normal data can be used, for each normal image $x_n$, we directly obtain its patched feature $q_n$:

\begin{equation}
    q_n=F_{\Phi}(x_n)
    \label{eq:1}
\end{equation}

As the detection system runs, some annotated anomalies become accessible and can be incorporated into the training of DMAD. For each seen anomaly $x_{a_s}$, we additionally employ a $Filter$ operation to isolate the anomalous parts from its extracted patch features $F_{\Phi}(x_{a_s})$. $Filter$ is adopted because we have observed that image defects generally constitute a small portion of the image, which needs to be filtered out. We conducted ablation experiments in \Cref{sec:ablation} to illustrate the importance of the $Filter$. More specifically, we denoted the annotation of a defective image as $y \in \{0,1\}^{1 \times H \times W}$. If the position $(h,w)$ in the image is normal, $y(h,w)=0$; otherwise, $y(h,w)=1$. We initially use bilinear interpolation to scale $y$ to match the resolution of the feature, and then filtered out the corresponding anomalous parts. Therefore, for each defective image $x_a$, we could calculate its anomalous patch feature $q_a \in \mathbb{R}^{N_f \times C}$:

\begin{equation}
    q_a=Filter(F_{\Phi}(x_{a_s}),y)
    \label{eq:2}
\end{equation}

Here, $N_f$ represents the number of anomalous patches, and $N_f \ll N$. The patch features will subsequently be enhanced by dual memory bank.

\label{sec:PFE}

% ---------------------------------------
% Dual Memory Bank-based Knowledge Enhancement
% ---------------------------------------
\subsection{Dual Memory Bank-based Knowledge Enhancement}
\begin{figure}
    \centering
    \includegraphics[width=1\textwidth]{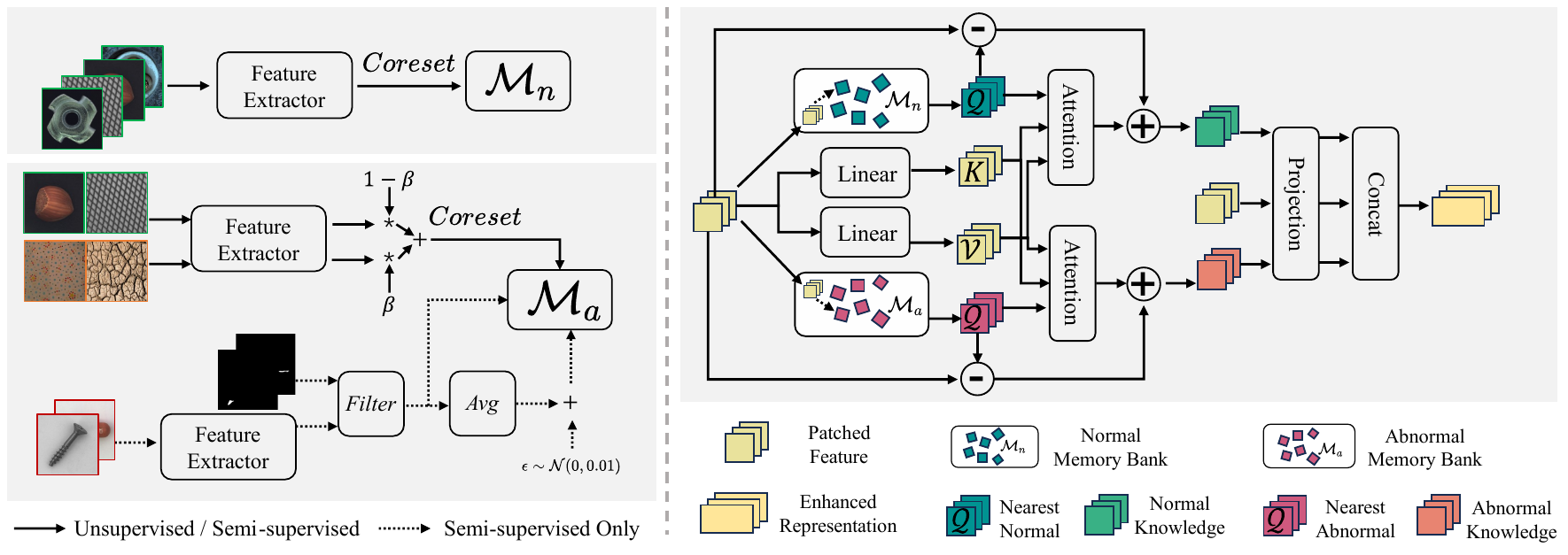}
    \caption{ An intuitive illustration of the Dual Memory Bank's construction is depicted on the \textbf{left}. A schematic diagram of the Knowledge Enhancement based on the Dual Memory Bank is shown on the \textbf{right}. DMAD uses the Dual Memory Bank to create an enhanced representation encompassing normal and abnormal knowledge.}
  \label{fig:framework_detail}
\end{figure}

To learn a comprehensive, unified decision boundary, it is more effective for the model to consider the entire training set, taking into account all different objects simultaneously. Therefore, we employ a dual memory bank $\mathcal{M_D}$ to introduce additional useful knowledge for anomaly detection. $\mathcal{M_D}$ comprises a normal memory bank, denoted as $\mathcal{M}_n$, which stores normal patterns, and an abnormal memory bank, denoted as $\mathcal{M}_a$, which stores all potential defect patterns. First, we will introduce the construction of the $\mathcal{M_D}$ (\Cref{fig:framework_detail}. Left). Following this, we will explain how to utilize $\mathcal{M_D}$ to extract additional knowledge, and subsequently use this knowledge to acquire an enhanced representation (\Cref{fig:framework_detail}. Right).

\noindent\textbf{Dual Memory Bank Construction.}
To construct $\mathcal{M}_n$, the patched features of all normal data $\{\mathcal{X}_n^i\}_{i=1}^{M}$ are extracted and then processed using a coreset sampling algorithm $Coreset$ to retain 2\% of them:

\begin{equation}
    \mathcal{M}_n=Coreset(\underset{{x_n \in \{\mathcal{X}_n^i\}_{i=1}^{M}}}{\bigcup}F_{\Phi}(x_n))
    \label{eq:3}
\end{equation}

The construction of $\mathcal{M}_a$ is tailored to the specific experimental setting. For a general unified setting, annotated anomalies are not available. To address this, we introduce randomly sampled outlier data $x_o \in \mathbb{R}^{3\times H\times W}$ from DTD\cite{cimpoi2014describing} dataset $\mathcal{X}_o$ to construct the $\mathcal{M}_a$, the ablation study presented in \Cref{sec:ablation} demonstrates the benefits of this. DTD dataset contains diverse textures, amalgamated with the normal data is a efficient strategy to construct the pseudo anomalies. More specifically, we use a $\beta$ parameter to fuse $x_o$ with a randomly sampled normal image $x_n$ at the feature-level, thereby creating the possible defects set $\mathcal{M}_o$:

\begin{equation}
    \begin{aligned}
    \mathcal{M}_o=Coreset(\underset{x_o \in \mathcal{X}_o, x_n \in \{\mathcal{X}_n^i\}_{i=1}^{M}}{\bigcup}(\beta F_{\Phi}(x_o) + (1-\beta)F_{\Phi}(x_n)))
    \end{aligned}
    \label{eq:4}
\end{equation}

In this context, we simply set $\beta=0.6$. In a general unified setting, $\mathcal{M}_a$ is essentially $\mathcal{M}_o$. Once annotated anomalies $\{\mathcal{X}_{a_s}^i\}_{i=1}^{M}$ become available, we are able to compute the filtered anomalous patch feature set of observed annotated anomalies, denoted as $\mathcal{M}_{a_s}$:

\begin{equation}
    \mathcal{M}_{a_s} = \underset{x_{a_s} \in \{\mathcal{X}_{a_s}^i\}_{i=1}^{M}, y \in \mathcal{Y}}{\bigcup}Filter(F_{\Phi}(x_{a_s}),y)
    \label{eq:5}
\end{equation}

Given the extreme data imbalance between normal and abnormal instances, training the model presents a significant challenge. Based on an intuitively assumption, that in a well-trained embedding space, the data points surrounding the abnormal feature are also likely to be abnormal. Thus we propose the \textbf{Anomaly Center Sampling} strategy, that is adding a small perturbation $\epsilon \sim \mathcal{N}(0,0.01)$ to the calculated average abnormal feature $q_a^{\text{avg}} \in \mathbb{R}^{N \times C}$ to generate pseudo abnormal features set $\mathcal{M}_p$.

For the unified semi-supervised setting, the abnormal memory bank is actually a union of $\mathcal{M}_o$, $\mathcal{M}_{a_s}$ and $\mathcal{M}_p$. Once the $\mathcal{M}_n$ and $\mathcal{M}_a$ constructed, they are saved on the disk and will unchanged. 

\noindent\textbf{Knowledge Enhancement.}
Under a general unified setting, for each patched normal feature $q_n$ that we obtain from \Cref{eq:1}, we identify its nearest neighbour from $\mathcal{M}_n$ and $\mathcal{M}_{a}$ respectively, denoted as $q_{n \cdot n}$ and $q_{n \cdot a}$:

\begin{equation}
    \begin{aligned}
        q_{n \cdot n}=\underset{q^* \subset \mathcal{M}_{n}}{\arg \min}||q_n-q^*|| \\
        q_{n \cdot a}=\underset{q^* \subset \mathcal{M}_{a}}{\arg \min}||q_n-q^*||
    \end{aligned}
    \label{eq:7}
\end{equation}

Subsequently, we compute a distance $dist$ between the feature and its two nearest neighbour features, denoted as $d_{n \cdot n}$ and $d_{n \cdot a}$ separately:

\begin{equation}
    \begin{aligned}
    d_{n \cdot n}=q_n-q_{n,n} \\
    d_{n \cdot a}=q_n-q_{n,a}
    \end{aligned}
    \label{eq:8}
\end{equation}

Moreover, we further utilize cross-attention to calculate the attention between the feature and its two nearest neighbour features. We consider the two nearest neighbour features $q_{n \cdot n}$ and $q_{n \cdot a}$ as the query embedding $\mathcal{Q}_n$ and $\mathcal{Q}_a$, and use a linear layer to embed the $q_n$ into key $\mathcal{K}_n \in \mathbb{R}^{N \times C}$ and value $\mathcal{V}_n \in \mathbb{R}^{N \times C}$. The two attention matrix $\mathcal{A}ttn$ is calculated as:

\begin{equation}
    \begin{aligned}
        \mathcal{A}_n=\text{softmax}(\mathcal{Q}_n(\mathcal{K}_n)^T\mathcal{V}_n) \\
        \mathcal{A}_a=\text{softmax}(\mathcal{Q}_a(\mathcal{K}_n)^T\mathcal{V}_n)
    \end{aligned}
    \label{eq:9}
\end{equation}

Hence, the knowledge $k \in \mathbb{R}^{N \times C}$ derived from the dual memory bank is denoted as the addition of $dist$ and $\mathcal{A}ttn$:

\begin{equation}
    \begin{aligned}
        k_{n \cdot n}=d_{n \cdot n}+\mathcal{A}_n \\
        k_{n \cdot a}=d_{n \cdot a}+\mathcal{A}_a
    \end{aligned}
    \label{eq:10}
\end{equation}

Both the distance and the attention matrix encapsulate the knowledge about the normality or abnormality of the feature. When scenarios transitions to unified semi-supervised setting, that is the anomalies is accessible, we perform the same operation to calculate the necessary knowledge $k_{a \cdot n}$ and $k_{a \cdot a}$ of the anomalous feature $q_a$ of each seen anomaly $x_{a_s}$. It's important to note that in unified semi-supervised setting, the attention matrix is not utilized. This is because we discovered that calculating distances independently yields superior results. The ablation study of this part is shown in \Cref{tab:ablationMain}.

After acquiring the knowledge, a projection layer $\mathcal{P}_\theta$ is applied to the feature and two parts of knowledge. Subsequently, the feature itself with two parts of knowledge are combined to form enhanced normal representation $o_n \in \mathbb{R}^{N \times 3C }$:

\begin{equation}
    o_n=Cat(\mathcal{P}_{\theta}(q_{n}),\mathcal{P}_{\theta}(k_{n \cdot n}),\mathcal{P}_{\theta}(k_{n \cdot a}))
    \label{eq:12}
\end{equation}

$Cat$ denotes the concatenation operation. For unified semi-supervised scenarios, we also have enhanced abnormal representation $o_a \in \mathbb{R}^{N_f \times 3C }$:
\begin{equation}
    o_a=Cat(\mathcal{P}_{\theta}(q_{a}),\mathcal{P}_{\theta}(k_{a \cdot n}),\mathcal{P}_{\theta}(k_{a \cdot a}))
    \label{eq:12}
\end{equation}

\label{sec:DMBKE}

\subsection{Anomaly Score Mapper}
We employ a MLP, denoted as $\Psi$, to learn a mapping between our constructed enhanced representation $o$, and the anomaly score $S \in \mathbb{R}^{H_0 \times W_o}$. A hinge loss function is utilized to optimize the network. In a general unified (multi-class) scenarios, where no negative samples are available, we adopt a feature augmentation strategy~\cite{liu2023simplenet}. This strategy involves using a Gaussian noise generator to create pseudo negative samples $o_p \in \mathbb{R}^{N \times 3C}$. When annotated anomalies are accessible, a three parts hinge loss is used for the model’s optimization:

\begin{equation}
    \begin{aligned}
    \mathcal{L} = & \max(0, 0.5 - \Psi(o_n)) \\
    & + \lambda_1 \max(0, 0.5 + \Psi(o_p)) \\
    & + \lambda_2 \max(0, 0.5 + \Psi(o_a)).
    \end{aligned}
    \label{eq:13}
\end{equation}

In the general unified setting, $\lambda_1=1,\lambda_2=0$, while in the unified semi-supervised setting, $\lambda_1=0.5,\lambda_2=15$.

\label{sec:ASP}

% ---------------------------------------
% Anomaly Detection and Localization
% ---------------------------------------
\subsection{Anomaly Detection and Localization}
Given a test image $x_{\text{test}}$, we can utilize a well-trained DMAD to derive its patch-level anomaly scores $S_{\text{test}}$. We adopt the average of the top-5 anomaly scores as its image-level score. For the pixel-level score, we initially apply bilinear interpolation to $S_{\text{test}}$, follow up with a Gaussian smoothing to refine its values.

\section{Experiments}
\subsection{Datasets and Evaluation Metrics}

\noindent\textbf{MVTec-AD.} MVTec-AD~\cite{bergmann2019mvtec}, a widely recognized anomaly detection benchmark, encompasses a diverse dataset of 5,354 high-resolution images from various domains. This dataset is categorized into 5 types of textures and 10 types of objects. The data is divided into training and testing sets, with the training set containing 3,629 anomaly-free images, ensuring a focus on normal samples. On the other hand, the test set consists of 1,725 images, providing a mix of both normal and abnormal samples for comprehensive evaluation. To aid in the anomaly localization evaluation, pixel-level annotations are provided.

\noindent\textbf{VisA.} 
The VisA~\cite{zou2022spot} dataset comprises 10,821 high-resolution images, including 9,621 normal images and 1,200 anomaly images. This dataset is organized into 12 unique object classes. These 12 object classes can be further categorized into three distinct object types: Complex Structures, Multiple Instances, and Single Instances. For our study, we specifically utilized the \textbf{2-class few-shot} setting split of the VisA dataset, where both the training and testing sets contain normal and anomalous samples.

\noindent\textbf{Evaluation Metrics.}
We utilize a comprehensive suite of standard evaluation metrics, which includes the Area Under the Receiver Operating Characteristic Curve (AUROC), Average Precision (AP), and the maximum F1-score (F1max). Additionally, for anomaly localization, we employ the Per-Region-Overlap (PRO) metric.
\begin{table}[t]
    \caption{Image-level AUROC/AP results on the MVTec-AD and VisA datasets for both unsupervised and semi-supervised scenarios under a unified (multi-class) setting.}
    \label{tab:imgRes}
    \centering
    \setlength{\tabcolsep}{3pt}
    \scalebox{0.47}{
        \begin{tabular}{l | c c c | c c c | c c c | c c c}
            \hline
            \multirow{2}{*}{\textbf{Datasets}} & \multicolumn{3}{c|}{\textbf{Unsupervised}} & \multicolumn{3}{c|}{\textbf{1 Anomaly}} & \multicolumn{3}{c|}{\textbf{5 Anomalies}} & \multicolumn{3}{c}{\textbf{10 Anomalies}} \\ \cline{2-13}
            & \textbf{UniAD}~\cite{you2022unified} & \textbf{SimpleNet$^*$}~\cite{liu2023simplenet} & \textbf{DMAD} & \textbf{DRA$^*$}~\cite{ding2022catching} & \textbf{BGAD$^*$}~\cite{yao2023explicit} & \textbf{DMAD} & \textbf{DRA$^*$}~\cite{ding2022catching} & \textbf{BGAD$^*$}~\cite{yao2023explicit} & \textbf{DMAD} & \textbf{DRA$^*$}~\cite{ding2022catching} & \textbf{BGAD$^*$}~\cite{yao2023explicit} & \textbf{DMAD} \\ \hline \hline
            Carpet     & \textbf{99.8} / \textbf{99.8} & 93.6 / 98.1 & 93.7 / 98.1 & 98.3 / 98.3 & \textbf{99.2} / 99.6 & 99.0 / \textbf{99.7} & 99.2 / \textbf{99.8} & \textbf{99.5} / \textbf{99.8} & 99.1 / 99.7 & 98.4 / 99.5 & \textbf{99.6} / \textbf{99.7} & 98.5 / 99.5 \\ %\hline
            Grid       & \textbf{98.2} / 97.5 & 97.9 / \textbf{99.4} & 96.8 / 99.0 & 90.4 / 93.4 & 96.6 / 97.2 & \textbf{97.5} / \textbf{99.2} & 96.2 / 98.7 & 97.6 / 98.8 & \textbf{98.1} / \textbf{99.4} & 83.6 / 93.5 & 98.1 / 97.2 & \textbf{99.5} / 99.8 \\ %\hline
            Leather    & \textbf{100} / \textbf{100} & \textbf{100} / \textbf{100} & \textbf{100} / \textbf{100} & 98.4 / 99.3 & \textbf{100} / \textbf{100} & \textbf{100} / \textbf{100} & 98.9 / 99.7 & \textbf{100} / \textbf{100} & \textbf{100} / \textbf{100} & \textbf{100} / \textbf{100} & \textbf{100} / \textbf{100} & \textbf{100} / \textbf{100} \\ %\hline
            Tile       & 99.3 / 99.7 & 99.5 / 99.8 & \textbf{100} / \textbf{100} & 95.5 / 98.1 & 99.9 / \textbf{100} & \textbf{100} / \textbf{100} & 97.2 / 99.1 & \textbf{100} / \textbf{100} & \textbf{100} / \textbf{100} & 98.0 / 99.2 & \textbf{100} / \textbf{100} & \textbf{100} / \textbf{100} \\ %\hline
            Wood       & 98.6 / 99.5 & 99.5 / 99.8 & \textbf{99.7} / \textbf{99.9} & 98.1 / 99.5 & 96.3 / 97.5 & \textbf{99.4} / \textbf{99.8} & \textbf{99.8} / \textbf{99.9} & 98.3 / 99.3 & \textbf{99.8} / \textbf{99.9} & 99.5 / \textbf{99.8} & 99.0 / 99.6 & \textbf{99.6} / \textbf{99.8} \\ 
            Bottle     & 99.7 / \textbf{100} & \textbf{100} / \textbf{100} & \textbf{100} / \textbf{100} & \textbf{100} / 99.6 & 99.8 / 99.7 & \textbf{100} / \textbf{100} & 99.7 / 99.9 & \textbf{100} / \textbf{100} & \textbf{100} / \textbf{100} & 99.2 / 99.7 & \textbf{100} / \textbf{100} & \textbf{100} / \textbf{100} \\ %\hline
            Capsule    & 86.9 / 96.7 & 84.2 / 96.5 & \textbf{89.8} / \textbf{97.8} & 72.0 / 95.0 & 87.6 / 94.3 & \textbf{93.7} / \textbf{98.7} & 73.4 / 91.6 & 91.4 / 96.8 & \textbf{96.1} / \textbf{99.2} & 80.1 / 86.8 & 93.4 / 97.0 & \textbf{97.9} / \textbf{99.6} \\ %\hline
            Pill       & \textbf{93.7} / \textbf{98.6} & 88.9 / 97.7 & 91.8 / 98.5 & 73.6 / 91.4 & 83.2 / 94.7 & \textbf{93.2} / \textbf{98.6} & 79.3 / 95.7 & 87.4 / 96.7 & \textbf{92.9} / \textbf{99.1} & 76.4 / 94.9 & 92.3 / 97.7 & \textbf{94.0} / \textbf{98.8} \\ %\hline
            Transistor & \textbf{99.8} / \textbf{99.2} & 96.4 / 95.6 & 98.8 / 98.2 & 64.4 / 62.8 & 93.0 / 88.3 & \textbf{98.6} / \textbf{98.0} & 64.7 / 53.4 & 99.2 / 96.9 & \textbf{99.6} / \textbf{99.3} & 66.3 / 60.7 & 98.2 / 96.6 & \textbf{99.8} / \textbf{99.6} \\ %\hline
            Zipper     & 95.8 / 98.3 & \textbf{99.6} / \textbf{99.9} & 99.5 / \textbf{99.9} & 95.7 / 96.0 & 98.2 / 99.5 & \textbf{99.7} / \textbf{99.9} & 96.6 / 99.1 & \textbf{99.6} / \textbf{99.9} & 99.4 / 99.8 & 98.1 / 99.5 & 99.9 / \textbf{100} & \textbf{100} / \textbf{100} \\ %\hline
            Cable      & 95.2 / 95.6 & 98.1 / 98.8 & \textbf{99.0} / \textbf{99.4} & 74.2 / 92.3 & 93.1 / 92.1 & \textbf{98.4} / \textbf{99.1} & 89.9 / 94.0 & 93.1 / 94.4 & \textbf{98.3} / \textbf{98.9} & 80.1 / 88.9 & 97.1 / 95.7 & \textbf{99.7} / \textbf{99.8} \\ %\hline
            Hazelnut   & 99.8 / 99.9 & 99.9 / 99.9 & \textbf{100} / \textbf{100} & 95.0 / 97.9 & \textbf{100} / 99.8 & \textbf{100} / \textbf{100} & 89.9 / 95.4 & \textbf{100} / 99.9 & \textbf{100} / \textbf{100} & 98.7 / 99.3 & 99.8 / 99.7 & \textbf{100} / \textbf{100} \\ %\hline
            Metal\_Nut  & \textbf{99.2} / 99.7 & 98.1 / 99.5 & 99.0 / \textbf{99.8} & 92.0 / 98.0 & 99.3 / 99.2 & \textbf{99.4} / \textbf{99.9} & 98.5 / 99.6 & 99.3 / 99.7 & \textbf{100} / \textbf{100} & 97.7 / 99.4 & 99.1 / 99.5 & \textbf{99.3} / \textbf{99.8} \\ %\hline
            Screw      & \textbf{87.5} / \textbf{94.9} & 77.2 / 91.6 & 84.1 / 93.7 & 76.7 / 94.6 & 84.5 / 92.8 & \textbf{90.2} / \textbf{96.3} & 87.6 / 95.4 & 88.1 / 91.5 & \textbf{92.9} / \textbf{97.5} & 90.0 / 94.9 & 84.2 / 92.1 & \textbf{97.3} / \textbf{99.0} \\
            Toothbrush & \textbf{94.2} / \textbf{96.6} & 89.7 / 95.4 & 91.7 / 96.4 & 69.0 / 90.7 & \textbf{96.0} / \textbf{97.8} & 92.5 / 96.8 & 78.0 / 88.3 & \textbf{99.0} / 97.9 & 95.7 / \textbf{98.0} & 75.4 / 87.4 & 96.7 / \textbf{96.9} & \textbf{99.2} / \textbf{99.5} \\ 
            \hline
            \textbf{MVTec-AD} & \textbf{96.5} / 98.4 & 94.8 / 98.1 & 96.3 / \textbf{98.7} & 86.2 / 93.8 & 95.1 / 96.8 & \textbf{97.4} / \textbf{99.1} & 89.9 / 99.1 & 96.8 / 98.1 & \textbf{98.3} / \textbf{99.4} & 88.4 / 93.6 & 97.1 / 98.1 & \textbf{99.0} / \textbf{99.7} \\ \hline
            Candle     & 85.4 / 20.3 & 94.1 / 82.5 & \textbf{95.6} / \textbf{83.1}  & 92.2 / 69.7 & \textbf{94.6} / 68.8 & \textbf{94.6} / \textbf{84.9} & 90.2 / 69.6 & 97.3 / 88.8 & \textbf{97.7} / \textbf{90.1} & 90.5 / 63.6 & 97.5 / 85.3 & \textbf{98.3} / \textbf{92.4} \\ %\hline
            Capsules    & 57.3 / 15.3 & 68.5 / 35.2 & \textbf{72.8} / \textbf{39.8} & 60.1 / 22.3 & 71.3 / 33.0 & \textbf{71.7} / \textbf{40.5} & 77.5 / 37.1 & 76.0 / \textbf{56.2} & \textbf{79.0} / 53.9 & 76.4 / 35.9 & \textbf{80.9} / \textbf{63.5} & 79.4 / 58.0 \\ %\hline
            Cashew       & 88.7 / 58.2 & 93.6 / 81.0 & \textbf{93.8} / \textbf{81.9} & 95.7 / 86.1 & \textbf{96.3} / 84.8 & 95.7 / \textbf{87.4} & \textbf{96.2} / 85.6 & 94.1 / \textbf{85.8} & 95.1 / 85.6 & \textbf{97.4} / \textbf{91.1} & 94.4 / 84.5 & 96.9 / 90.7 \\ %\hline
            Chewinggum    & \textbf{97.8} / 93.3 & 97.0 / \textbf{95.9} & 97.1 / 95.2 & 93.6 / 88.9 & \textbf{98.9} / 96.4 & 98.3 / \textbf{96.7} & 96.4 / 92.9 & \textbf{98.9} / 96.6 & 98.1 / \textbf{96.9} & 96.4 / 94.1 & \textbf{99.0} / \textbf{97.6} & 98.7 / \textbf{97.6} \\ %\hline
            Fryum       & 81.3 / 43.3 & 87.3 / 76.2 & \textbf{90.9} / \textbf{78.6} & 77.5 / 63.0 & 90.3 / 72.3 & \textbf{92.8} / \textbf{83.4} & 90.5 / 75.4 & \textbf{93.7} / \textbf{83.1} & 93.4 / 83.0 & 88.4 / 73.6 & 93.3 / 80.3 & \textbf{96.5} / \textbf{89.6} \\ %\hline
            Macaroni1       & 77.1 / 15.8 & \textbf{87.3} / 46.6 & 86.9 / \textbf{55.9} & 69.2 / 26.0 & \textbf{90.0} / 53.4 & 87.2 / \textbf{56.8} & 93.1 / 73.5 & \textbf{96.0} / \textbf{77.4} & 93.7 / 75.1 & 91.0 / 67.1 & 94.7 / 74.8 & \textbf{97.3} / \textbf{87.4} \\ %\hline
            Macaroni2 & 72.4 / 15.0 & 69.5 / 22.8 & \textbf{72.6} / \textbf{24.8} & 53.5 / 11.2 & \textbf{74.9} / 20.7 & 73.7 / \textbf{31.7} & 82.0 / \textbf{57.0} & \textbf{82.4} / 37.0 & 78.8 / 37.4 & 83.4 / \textbf{61.4} & 85.2 / 45.7 & \textbf{85.4} / 57.5 \\ %\hline
            Pcb1     & 90.3 / 45.9 & \textbf{95.6} / \textbf{78.5} & 94.9 / 77.0 & 87.1 / 54.8 & 89.8 / 47.5 & \textbf{95.1} / \textbf{77.7} & 70.7 / 48.6 & 92.2 / 56.1 & \textbf{97.7} / \textbf{86.7} & 83.3 / 52.2 & 94.3 / 67.9 & \textbf{98.8} / \textbf{91.8} \\ %\hline
            Pcb2      & 87.1 / 48.2 & 93.8 / \textbf{79.7} & \textbf{94.1} / 77.5 & 85.8 / 49.0 & 84.7 / 56.9 & \textbf{97.5} / \textbf{87.8} & 83.5 / 51.9 & 88.9 / 61.1 & \textbf{98.2} / \textbf{90.7} & 75.6 / 23.4 & 92.7 / 75.0 & \textbf{98.5} / \textbf{92.5} \\ %\hline
            Pcb3     & 80.1 / 30.5 & 84.9 / 58.2 & \textbf{87.4} / \textbf{62.4} & 77.1 / 24.6 & 89.7 / 58.0 & \textbf{89.9} / \textbf{69.5} & 68.4 / 22.6 & \textbf{92.3} / 68.8 & 90.7 / \textbf{71.0} & 72.8 / 17.9 & \textbf{93.7} / \textbf{76.2} & 91.7 / 71.3 \\ %\hline
            Pcb4   & 97.1 / 68.1 & 97.9 / \textbf{91.2} & \textbf{98.5} / 90.3 & 75.4 / 33.7 & 97.1 / 74.9 & \textbf{99.0} / \textbf{94.6} & 86.7 / 74.6 & \textbf{98.8} / 87.2 & 98.7 / \textbf{95.4} & 80.7 / 63.8 & 99.3 / 92.2 & \textbf{99.7} / \textbf{97.2} \\ %\hline
            Pipe\_Fryum  & 91.2 / 61.4 & 91.0 / \textbf{79.9} & \textbf{91.6} / \textbf{79.9} & 92.1 / 83.1 & \textbf{98.5} / \textbf{95.7} & 95.7 / 85.7 & \textbf{97.8} / 94.3 & 97.2 / 93.1 & 96.1 / 91.7 & 97.6 / 93.2 & \textbf{98.9} / \textbf{94.8} & 98.1 / 93.0 \\ \hline
            \textbf{VisA} & 83.9 / 42.9 & 88.0 / 69.0 & \textbf{89.7} / \textbf{70.5} & 80.0 / 51.0 & 89.7 / 63.5 & \textbf{90.9} / \textbf{74.7} & 86.9 / 65.2 & 92.3 / 74.3 & \textbf{93.1} / \textbf{79.8} & 86.1 / 61.4 & 93.7 / 78.2 & \textbf{94.9} / \textbf{84.9} \\ \hline
        \end{tabular}
    }
\end{table}

\begin{table}[]
\caption{Results of F1max and PRO comparison on VisA datasets. }
\label{tab:moreRes}
\setlength{\tabcolsep}{4.5pt}
\centering
\begin{tabular}{c|c|c|c|c|c}
\hline
\multirow{2}{*}{} & \multicolumn{3}{c|}{Unsupervised} & \multicolumn{2}{c}{10 Anomalies} \\ \cline{2-6} 
                  & \textbf{UniAD}~\cite{you2022unified} & \textbf{SimpleNet$^*$}~\cite{liu2023simplenet} & \textbf{DMAD} & \textbf{BGAD$^*$}~\cite{yao2023explicit} & \textbf{DMAD} \\ \hline
F1$\max$-Det    & 47.3 & 66.9 & \textbf{67.5} & 72.3 & \textbf{79.7} \\ \hline
F1$\max$-Loc    & 28.4 & 36.3 & \textbf{36.7} & \textbf{47.5} & 44.9 \\ \hline
PRO             & 83.0 & 85.6 & \textbf{85.7} & 91.9 & \textbf{92.9} \\ \hline
\end{tabular}
\end{table}
\vspace{-20pt}

\subsection{Implementation Details}
We utilize the WideResnet50~\cite{zagoruyko2016wide} as our pre-trained CNN backbone, extracting features from both the layer-2 and layer-3 to aggregate into a patched feature~\cite{roth2022towards}. The dimension of this feature is $1536$. For the projection layer, we employ a single fully connected layer to project the features and knowledge. The MLP consists of four nonlinear layers, each of which includes a linear layer, batch normalization, and leaky ReLU activation. Skip connections are applied between these nonlinear layers. The AdamW\cite{loshchilov2017decoupled} optimizer is used, with a learning rate of 0.0001 for the linear layer in the cross-attention and the projection layer, 0.0002 for the MLP. The weight decay for the MLP is set at 1e-5. The training process extends over 48 epochs, with a batch size of 32.

\subsection{Experiment Protocols}
In real-world scenarios, annotated anomalies may or may not be accessible, and the number of anomalies can fluctuate. To evaluate the performance of DMAD in real-world anomaly detection, we examine DMAD in both unsupervised and semi-supervised contexts within a unified setting. The details are as follows:
\begin{table}[t]
    \caption{Pixel-level AUROC/AP results on the MVTec-AD and VisA datasets for both unsupervised and semi-supervised scenarios under a unified (multi-class) setting.}
    \label{tab:pixRes}
    \centering
    \setlength{\tabcolsep}{5pt}
    \scalebox{0.5}{
        \begin{tabular}{l | c c c | c c | c c | c c}
            \hline
            \multirow{2}{*}{\textbf{Datasets}} & \multicolumn{3}{c|}{\textbf{Unsupervised}} & \multicolumn{2}{c|}{\textbf{1 Anomaly}} & \multicolumn{2}{c|}{\textbf{5 Anomalies}} & \multicolumn{2}{c}{\textbf{10 Anomalies}} \\ \cline{2-10}
            & \textbf{UniAD}~\cite{you2022unified} & \textbf{SimpleNet$^*$}~\cite{liu2023simplenet} & \textbf{DMAD} & \textbf{BGAD$^*$}~\cite{yao2023explicit} & \textbf{DMAD}&
            \textbf{BGAD$^*$}~\cite{yao2023explicit} & \textbf{DMAD}& \textbf{BGAD$^*$}~\cite{yao2023explicit} & \textbf{DMAD} \\ \hline \hline
            Carpet     & \textbf{98.5} / \textbf{51.7} & 96.5 / 38.4 & 97.5 / 42.3 & \textbf{99.2} / \textbf{69.1} & 98.8 / 60.8 & \textbf{99.4} / \textbf{72.1} & 99.1 / 71.0 & \textbf{99.4} / \textbf{75.1} & 99.1 / 72.3 \\ %\hline
            Grid       & 96.5 / 22.5 & 96.2 / 22.6 & \textbf{97.2} / \textbf{26.0} & \textbf{98.6} / \textbf{39.6} & 97.1 / 30.4 & \textbf{98.9} / 40.3 & 96.9 / \textbf{40.7} & \textbf{99.1} / \textbf{40.4} & 98.4 / 30.9 \\ %\hline
            Leather    & 98.8 / 34.1 & 98.7 / 31.9 & \textbf{98.9} / \textbf{35.6} & \textbf{99.6} / \textbf{59.3} & 99.0 / 35.9 & \textbf{99.7} / \textbf{61.1} & 99.0 / 38.5 & \textbf{99.7} / \textbf{66.0} & 98.5 / 29.0 \\ %\hline
            Tile       & 91.8 / 44.3 & \textbf{99.2} / 57.3 & 95.7 / \textbf{61.5} & 96.7 / 64.4 & \textbf{96.9} / \textbf{65.3} & \textbf{98.0} / \textbf{68.1} & 96.0 / 49.8 & \textbf{98.4} / \textbf{76.1} & 96.8 / 70.2 \\ %\hline
            Wood       & \textbf{93.2} / 37.9 & 91.5 / 36.7 & 92.9 / \textbf{41.7} & \textbf{96.7} / \textbf{59.3} & 93.3 / 36.2 & \textbf{96.9} / \textbf{61.2} & 92.3 / 48.8 & \textbf{97.2} / \textbf{60.4} & 94.9 / 44.6 \\
            Bottle     & 99.7 / \textbf{68.2} & 97.3 / 58.6 & 97.7 / 59.4 & 97.4 / 64.7 & \textbf{97.9} / \textbf{66.6} & \textbf{98.8} / \textbf{76.0} & 97.9 / 65.3 & \textbf{99.0} / \textbf{80.1} & 98.2 / 70.6 \\ %\hline
            Capsule    & 98.5 / \textbf{47.0} & 97.7 / 34.6 & \textbf{98.7} / 38.8 & 96.9 / 26.9 & \textbf{98.5} / \textbf{39.9} & 97.4 / 35.3 & \textbf{98.7} / \textbf{39.9} & 97.8 / 40.7 & \textbf{99.0} / \textbf{50.9} \\ %\hline
            Pill       & 95.0 / 40.9 & 96.5 / 72.6 & \textbf{97.2} / \textbf{74.6} & 95.1 / 38.4 & \textbf{97.2} / \textbf{77.4} & 95.6 / 43.0 & \textbf{98.8} / \textbf{76.4} & 98.6 / 74.5 & \textbf{98.9} / \textbf{85.4} \\ %\hline
            Transistor & \textbf{97.9} / \textbf{73.4} & 93.7 / 55.2 & 95.5 / 59.9 & 83.3 / 28.2 & \textbf{95.0} / \textbf{56.7} & 92.0 / 35.3 & \textbf{97.6} / \textbf{66.2} & 90.2 / 35.0 & \textbf{97.0} / \textbf{60.3} \\ %\hline
            Zipper     & 96.8 / 33.3 & 97.7 / \textbf{58.0} & \textbf{98.0} / 56.0 & \textbf{98.5} / 56.2 & 98.3 / \textbf{61.8} & \textbf{99.0} / \textbf{71.0} & 98.1 / 68.8 & \textbf{99.2} / 70.2 & \textbf{99.2} / \textbf{73.4} \\ %\hline
            Cable      & \textbf{97.3} / \textbf{53.2} & 96.3 / 50.5 & 96.9 / 48.7 & 85.4 / 31.7 & \textbf{96.6} / \textbf{47.1} & 92.5 / 42.2 & \textbf{97.4} / \textbf{49.9} & 92.0 / 38.1 & \textbf{98.1} / \textbf{56.4} \\ %\hline
            Hazelnut   & 98.1 / \textbf{53.8} & \textbf{98.2} / 50.9 & \textbf{98.2} / 51.2 & 97.8 / \textbf{62.7} & \textbf{98.2} / 52.3 & 98.1 / \textbf{62.1} & \textbf{98.7} / 57.9 & 98.6 / \textbf{62.9} & \textbf{98.7} / 60.3 \\ %\hline
            Metal\_nut  & 94.8 / 49.5 & 97.4 / 81.3 & \textbf{98.1} / \textbf{85.1} & 96.4 / 61.7 & \textbf{98.7} / \textbf{91.8} & 97.4 / 76.6 & \textbf{98.7} / \textbf{90.8} & 98.2 / 71.6 & \textbf{98.9} / \textbf{95.6} \\ %\hline
            Screw      & \textbf{98.3} / \textbf{25.0} & 95.9 / 18.0 & 97.4 / 23.8 & \textbf{98.6} / \textbf{32.5} & 98.1 / 23.4 & 98.7 / \textbf{30.1} & \textbf{98.8} / 27.8 & 98.5 / 30.5 & \textbf{99.1} / \textbf{30.8} \\ %\hline
            Toothbrush & \textbf{98.4} / 39.4 & 98.1 / 53.6 & 98.2 / \textbf{54.3} & 97.1 / 33.8 & \textbf{98.5} / \textbf{60.9} & 97.8 / 33.5 & \textbf{98.7} / \textbf{52.8} & \textbf{98.9} / 50.1 & \textbf{98.9} / \textbf{50.6} \\ \hline
            \textbf{MVTec-AD} & 96.8 / 44.9 & 96.5 / 48.0 & 97.2 / \textbf{50.6} & 95.8 / 48.6 & \textbf{97.5} / \textbf{53.8} & 97.3 / 53.9 & \textbf{97.7} / \textbf{56.4} & 97.6 / 58.1 & \textbf{98.2} / \textbf{58.7} \\ \hline
            Candle     & \textbf{99.2} / 6.30 & 98.6 / \textbf{14.4} & 97.6 / 11.7 & \textbf{99.5} / \textbf{19.7} & 98.2 / 12.8 & \textbf{99.6} / \textbf{25.7} & 99.3 / 12.4 & \textbf{99.7} / \textbf{30.1} & 98.3 / 18.8 \\ %\hline
            Capsules    & 84.3 / 0.30 & \textbf{97.1} / \textbf{15.1} & \textbf{97.1} / 12.8 & 97.7 / \textbf{18.4} & \textbf{98.0} / 16.8 & \textbf{98.9} / \textbf{50.7} & \textbf{98.9} / 42.7 & 99.2 / 56.7 & \textbf{99.4} / \textbf{59.2} \\ %\hline
            Cashew       & 99.6 / 54.6 & \textbf{99.7} / \textbf{74.3} & 99.3 / 69.2 & 95.5 / 21.1 & \textbf{99.7} / \textbf{69.9} & 95.8 / 27.4 & \textbf{99.7} / \textbf{69.6} & 96.3 / 33.7 & \textbf{99.6} / \textbf{66.2} \\ %\hline
            Chewinggum    & \textbf{99.2} / \textbf{58.3} & 98.9 / 29.3 & 98.8 / 31.6 & \textbf{99.9} / \textbf{69.4} & 99.1 / 57.5 & \textbf{99.9} / \textbf{80.1} & 99.2 / 62.6 & \textbf{99.9} / \textbf{78.8}  & 99.4 / 66.8 \\ %\hline
            Fryum       & \textbf{98.4} / 35.9 & 96.9 / 42.6 & 97.2 / \textbf{44.2} & 96.3 / 25.1 & \textbf{97.6} / \textbf{48.0} & 96.7 / 33.4 & \textbf{98.8} / \textbf{56.2} & 97.4 / 41.7 & \textbf{99.1} / \textbf{56.2} \\ %\hline
            Macaroni1       & \textbf{98.6} / 0.94 & 96.9 / 3.80 & 97.9 / \textbf{4.40} & \textbf{99.6} / \textbf{11.0} & 99.1 / 5.70 & \textbf{99.7} / \textbf{18.1} & 99.6 / 9.92 & \textbf{99.7} / \textbf{17.8} & \textbf{99.7} / 11.8 \\ %\hline
            Macaroni2 & 93.4 / 0.06 & 93.6 / 0.23 & \textbf{93.8} / \textbf{0.32} & \textbf{98.5} / \textbf{3.40} & 95.8 / 0.60 & \textbf{99.0} / \textbf{6.70} & 96.6 / 1.20 & \textbf{99.1} / \textbf{6.60} & 97.9 / 2.83 \\ %\hline
            Pcb1     & \textbf{99.5} / 42.1 & 99.2 / 80.9 & 99.3 / \textbf{82.3} & 98.8 / 53.5 & \textbf{99.7} / \textbf{73.4} & 98.7 / 60.6 & \textbf{99.6} / \textbf{81.1} & 99.1 / 83.9 & \textbf{99.7} / \textbf{84.1} \\ %\hline
            Pcb2      & \textbf{98.3} / 4.67 & \textbf{98.3} / \textbf{19.1} & \textbf{98.3} / 15.9 & 95.8 / \textbf{19.0} & \textbf{99.0} / 18.1 & 96.2 / \textbf{23.2} & \textbf{98.8} / 18.9 & 96.3 / \textbf{38.0} & \textbf{99.4} / 29.1 \\ %\hline
            Pcb3     & 98.1 / 9.41 & 98.3 / 12.1 & \textbf{98.8} / \textbf{19.8} & 96.0 / \textbf{20.7} & \textbf{98.7} / 19.7 & 96.3 / \textbf{30.1} & \textbf{99.0} / 27.3 & 96.5 / \textbf{34.9} & \textbf{99.2} / 25.0 \\ %\hline
            Pcb4   & \textbf{99.0} / 24.7 & 98.8 / \textbf{26.1} & 97.9 / 25.3 & 95.9 / 13.2 & \textbf{98.5} / \textbf{23.1} & 98.5 / \textbf{41.6} & \textbf{99.7} / 38.3 & 98.8 / \textbf{44.1} & \textbf{99.7} / 43.7 \\ %\hline
            Pipe\_fryum  & \textbf{99.7} / 49.5 & 99.6 / 62.1 & 99.5 / \textbf{64.2} & 98.8 / 36.9 & \textbf{99.6} / \textbf{57.4} & 99.4 / 51.7 & \textbf{99.7} / \textbf{63.8} & 99.5 / 55.4 & \textbf{99.7} / \textbf{61.2} \\ \hline
            \textbf{VisA} & 97.3 / 23.9 & 97.5 / 31.7 & \textbf{98.0} / \textbf{31.8} & 97.7 / 25.9 & \textbf{98.6} / \textbf{33.6} & 98.2 / 37.5 & \textbf{99.0} / \textbf{40.3} & 98.4 / 43.5 & \textbf{99.3} / \textbf{43.7} \\ \hline
        \end{tabular}
    }
\end{table}

\noindent\textbf{Unified Setting (Unsupervised).}
This is also known as a multi-class setting.  Following \cite{you2022unified}, the training data consists of normal images from all objects in the dataset, with no anomalies included. The test set contains both normal and abnormal images, and evaluation is conducted separately for each object.

\noindent\textbf{Unified Semi-Supervised Setting.}
This setting is applicable to real-world anomaly detection. To simulate varying quantities of anomalies that could be encountered in real-world scenarios, we set the possible number of accessible annotated anomalies for each object to 1, 5, and 10, respectively. The observed anomalies are randomly selected from each object's test set and are excluded during evaluation. It should be noted that for the VisA dataset, we directly select anomalies from its training set.
 % Additionally, we considered a special scenario where the obtainable anomalies are uniformly distributed, thus we manually selected 5 anomalies for each object, denoted as selected anomalies. 

\subsection{Anomaly Detection and Localization}
We conducted a series of both qualitative and quantitative comparison experiments on the MVTec-AD and VisA datasets. The results were subsequently analyzed. We selected the benchmark of the unified (multi-class) setting, UniAD~\cite{you2022unified}, a state-of-the-art one-class method, SimpleNet~\cite{liu2023simplenet}, and two semi-supervised methods, DRA~\cite{ding2022catching} and BGAD~\cite{yao2023explicit} for comparison. Due to the unavailability of the code for PRN~\cite{zhang2023prototypical}, we did not include it in our comparison. Furthermore, we did not present the pixel-level metric of DRA~\cite{ding2022catching}, as the code for its anomaly map generation has not been released yet. It should be noted that these methods do not provide results in a unified setting, so we implemented them ourselves, denoted by a $^*$ in the top right corner.

\begin{figure}[t]
    \centering
    \includegraphics[width=1\textwidth]{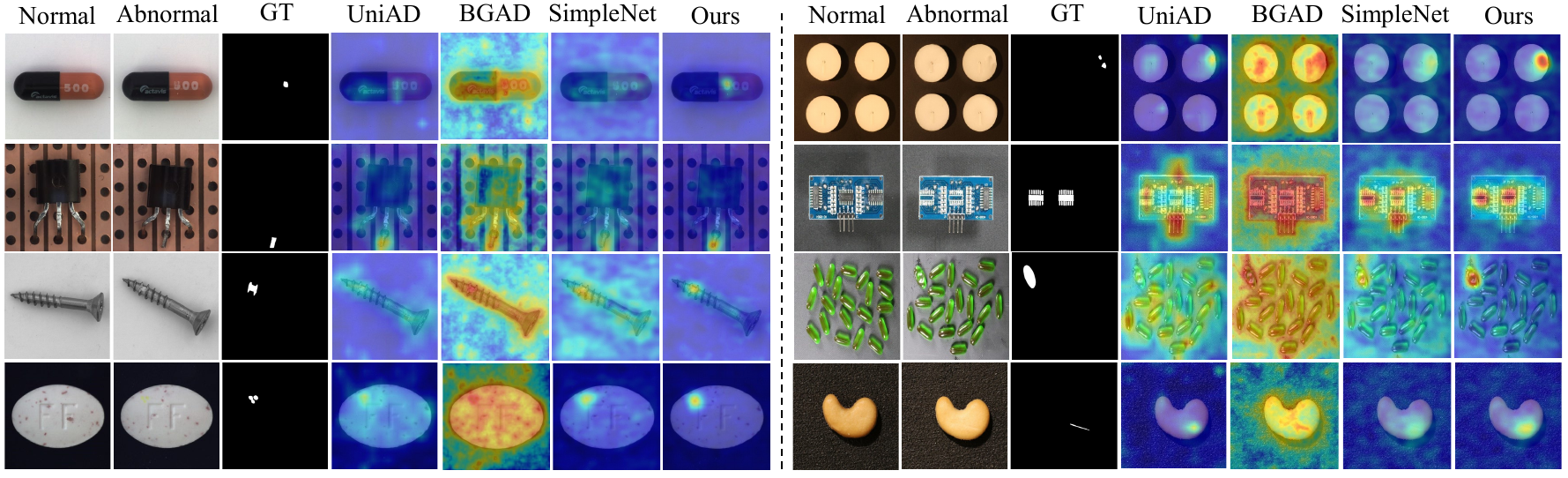}
    \caption{Qualitative comparison of our method with UniAD~\cite{you2022unified}, BGAD~\cite{yao2023explicit} and SimpleNet~\cite{liu2023simplenet} on MVTec-AD (Left) and VisA (Right) dataset.}
  \label{fig:visualization}
\end{figure}

\Cref{tab:imgRes} and \Cref{tab:moreRes} presents a quantitative comparison for anomaly detection. In unsupervised scenarios, DMAD, despite not being specifically designed for such situations, demonstrates performance on par with UniAD~\cite{you2022unified} when tested on the MVTec-AD dataset. Furthermore, it outperforms UniAD~\cite{you2022unified} on the VisA dataset, achieving gains of 5.8$\uparrow$ and 42.9$\uparrow$ in AUROC and AP, respectively. When a few annotated anomalies are available, DMAD utilizes the dual memory bank to learn a more precise decision boundary, thereby achieving state-of-the-art performance on both the MVTec-AD and VisA datasets. Specifically, in a setting with 10 anomalies, DMAD can achieve an AUROC of 99.0 and an AP of 99.7 for anomaly detection in MVTec-AD, and 94.9 AUROC, 79.7 F1$\max$ and 84.9 AP in VisA.
The anomaly localization performance comparison is shown in \Cref{tab:moreRes} and \Cref{tab:pixRes}. DMAD achieves the SOTA AUROC/AP/PRO metrics for all different settings, demonstrating its ability to effectively handle the challenges presented by real-world scenarios. More specifically, DMAD achieves up to 98.2 AUROC and 58.7 AP on the MVTec-AD dataset, as well as 99.3 AUROC, 43.7 AP and 92.9 PRO on the VisA dataset.

We also conduct a qualitative evaluation of anomaly localization, as shown in \Cref{fig:visualization}. Intuitively, our method facilitates a more accurate identification.

\subsection{Ablation Study}
\label{sec:ablation}

To validate the efficacy of the proposed modules, we conducted comprehensive ablation studies on the MVTec-AD dataset, as illustrated in \Cref{tab:ablationMain}. Det. and Loc. denote the results of Image-level and Pixel-level AUROC, respectively.

\noindent\textbf{Knowledge's Components.} DMAD utilizes a dual memory bank to calculate knowledge related to both normal and abnormal, incorporating a distance $dist$ and an attention matrix $\mathcal{A}ttn$. In unsupervised contexts, we observed that the application of $\mathcal{A}ttn$ boosts performance. Conversely, in semi-supervised contexts, we found that relying exclusively on $dist$ as the primary source of knowledge produces superior outcomes. We hypothesize that this discrepancy is attributable to the relatively sparse number of anomalies; an overemphasis on knowledge about these rare anomalies could potentially result in overfitting.

\begin{table}[t]
\caption{Ablation study results for our architecture.}
\label{tab:ablationMain}
\centering
\setlength{\tabcolsep}{5pt}
\begin{tabular}{c|c|c|c|c|c|c|c|c|c}
\hline
\multirow{3}{*}{} & \multirow{3}{*}{$Filter$} & \multicolumn{4}{c|}{Dual Memory Bank} & \multicolumn{2}{c|}{Knowledge} & \multirow{3}{*}{Det.} & \multirow{3}{*}{Loc.} \\ \cline{3-8}
   & & \multirow{2}{*}{$\mathcal{M}_n$} & \multicolumn{3}{c|}{$\mathcal{M}_a$} & \multirow{2}{*}{$dist$} & \multirow{2}{*}{$\mathcal{A}ttn$} & & \\ \cline{4-6}
   & & & $\mathcal{M}_o$ & $\mathcal{M}_{a_s}$ & $\mathcal{M}_p$ & & & & \\ \hline
\multirow{3}{*}{Unsupervised} &
   & $\checkmark$ &  &  &  & $\checkmark$ &  & 94.9 & 96.0 \\  
   & & $\checkmark$ & & & & $\checkmark$ & $\checkmark$ & 95.8 & 97.1 \\
   & & $\checkmark$ & $\checkmark$ & & & $\checkmark$ & $\checkmark$ & \textbf{96.3} & \textbf{97.2} \\ \hline
\multirow{4}{*}{Semi-supervised} &
   & $\checkmark$ & $\checkmark$ & $\checkmark$ & & $\checkmark$ & & 93.6 & 72.3 \\
   & $\checkmark$ & $\checkmark$ & $\checkmark$ & $\checkmark$ & & $\checkmark$ & & 98.7 & 98.1 \\
   & $\checkmark$ & $\checkmark$ & $\checkmark$ & $\checkmark$ & & $\checkmark$ & $\checkmark$ & 95.4 & 97.1\\
   & $\checkmark$ & $\checkmark$ & $\checkmark$ & $\checkmark$ & $\checkmark$ & $\checkmark$ &  & \textbf{99.0} & \textbf{98.2} \\ \hline
\end{tabular}
\vspace{-0.2cm}
\end{table}

\noindent\textbf{Abnormal Memory Bank's Components.} In unsupervised scenarios, no anomalies are available for use. Although a single normal memory bank could also be utilized for anomaly detection, we found that employing a pseudo abnormal memory bank $\mathcal{M}_o$, composed of pseudo abnormal features augmented with outlier data, could significantly enhance performance. This approach led to a 0.5$\uparrow$/0.1$\uparrow$ improvement in image-level AUROC and pixel-level AUROC. For semi-supervised scenarios, we proposed an anomaly center sampling strategy to generate an additional pseudo abnormal feature set. The integration of this set resulted in a noticeable enhancement in the model's performance, leading to a 0.3$\uparrow$/0.1$\uparrow$ improvement in image-level AUROC and pixel-level AUROC.

\noindent\textbf{Filter Operation.}
Introducing anomalies into the framework by directly adding a branch for seen anomalies $\mathcal{X}_{a_s}$ was ineffective, as anomalous images contain both normal and abnormal regions. Instead, we introduced a selective anomaly filter $Filter$, which significantly improved results.

\setlength{\tabcolsep}{2.5pt} % 调整单元格内的水平间距为5pt
\begin{table}[t]
    \caption{Quantitative comparison between different scenarios.}
    \label{tab:analysis}
    \centering
    \begin{tabular}{ c | c | c | c | c | c}
    \hline
    Class & Typical Defect & Similar Defect & Setting & Det. & Loc. \\ 
    \hline
    \multirow{3}{*}{macaroni1} & \multirow{3}{*}{\includegraphics[width=0.10\textwidth]{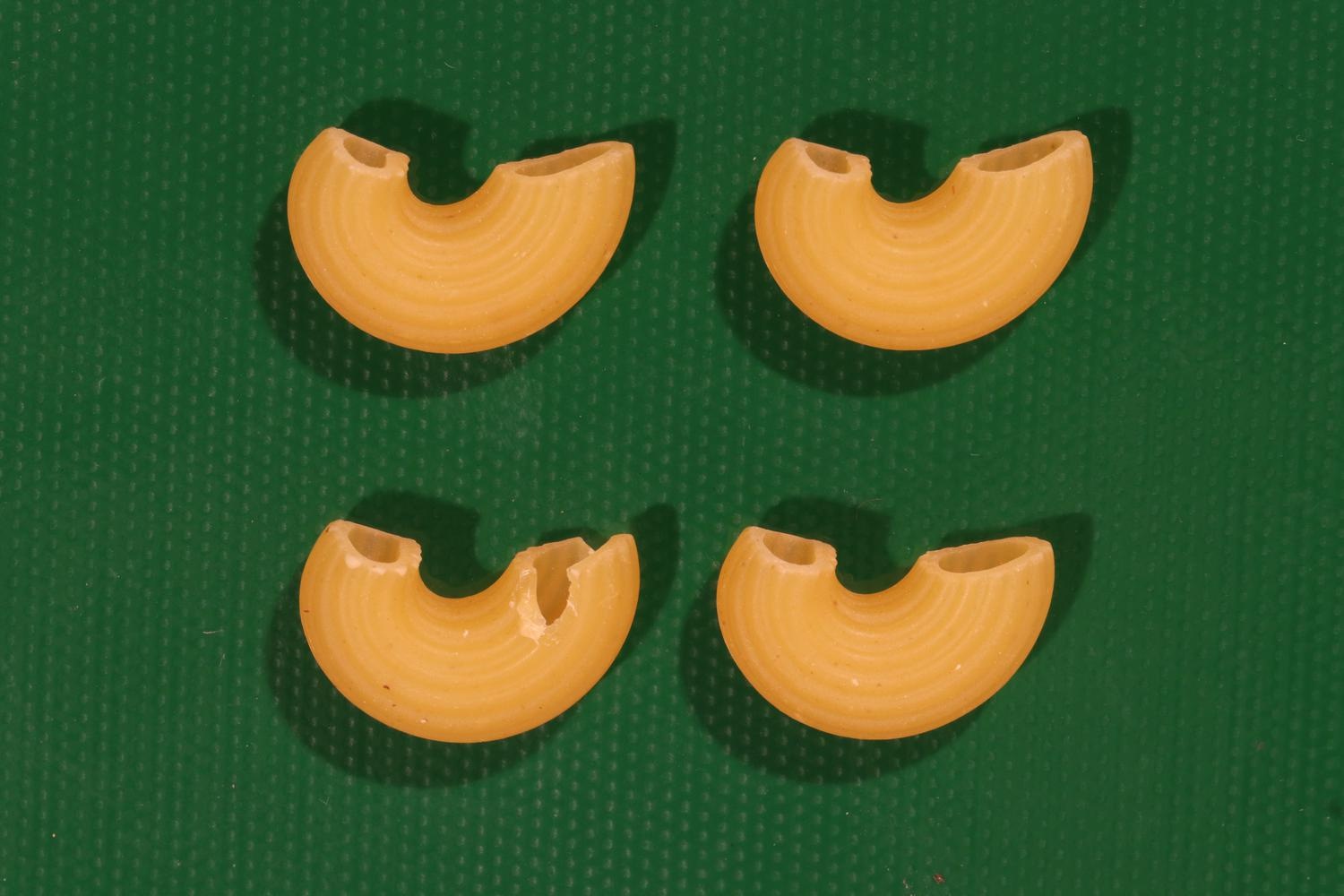}} & \multirow{3}{*}{\includegraphics[width=0.10\textwidth]{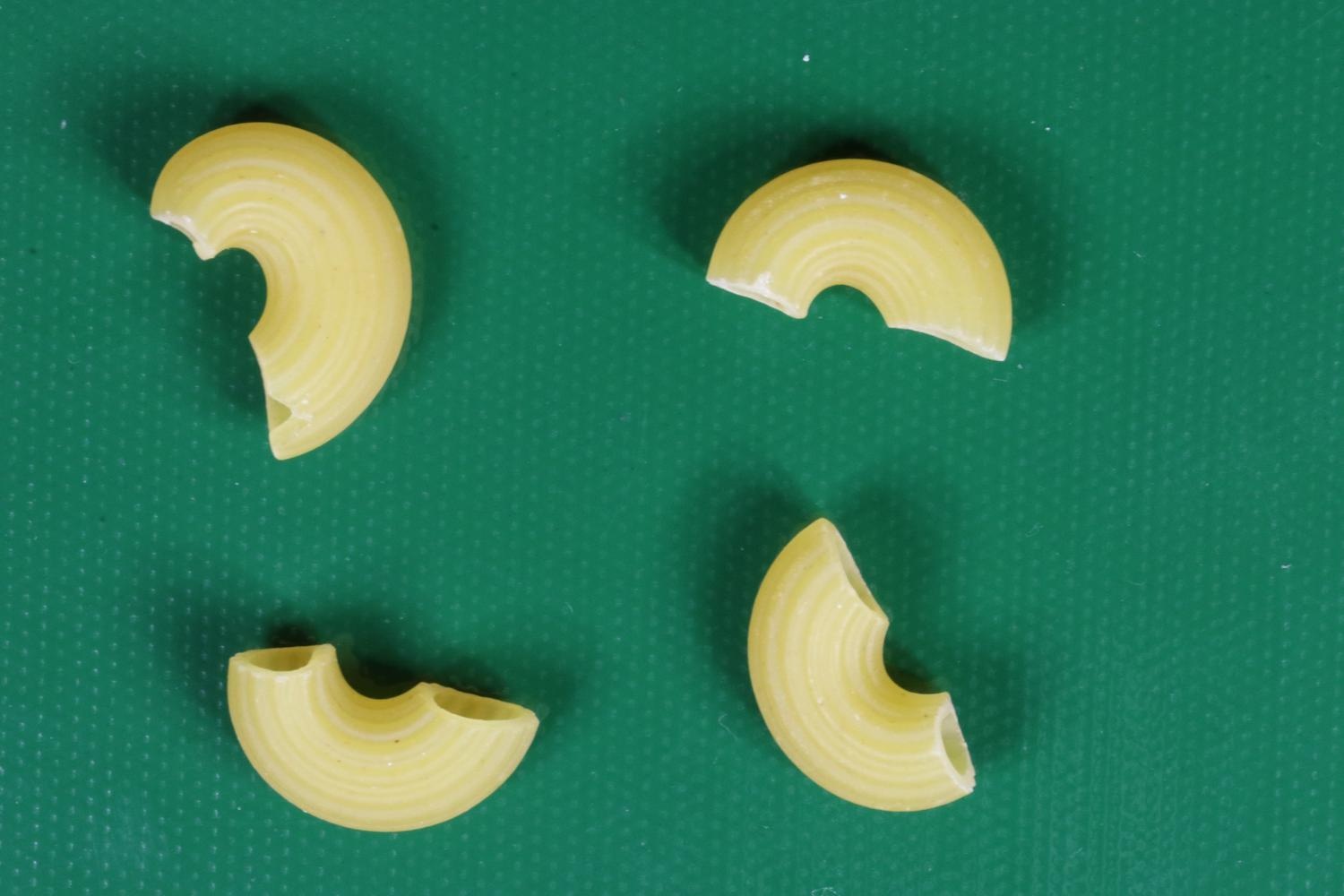}} & unsupervised & 86.9(+0.0) & 97.9(+0.0) \\ 
    &&& supervised$_{\text{others}}$ & 91.2(\textcolor{red}{+4.3}) & 99.2(\textcolor{red}{+1.2}) \\ 
    &&& supervised & 97.3(\textcolor{red}{+10.4}) & 99.7(\textcolor{red}{+1.7}) \\
    \hline
    \multirow{3}{*}{chewinggum} & \multirow{3}{*}{\includegraphics[width=0.08\textwidth]{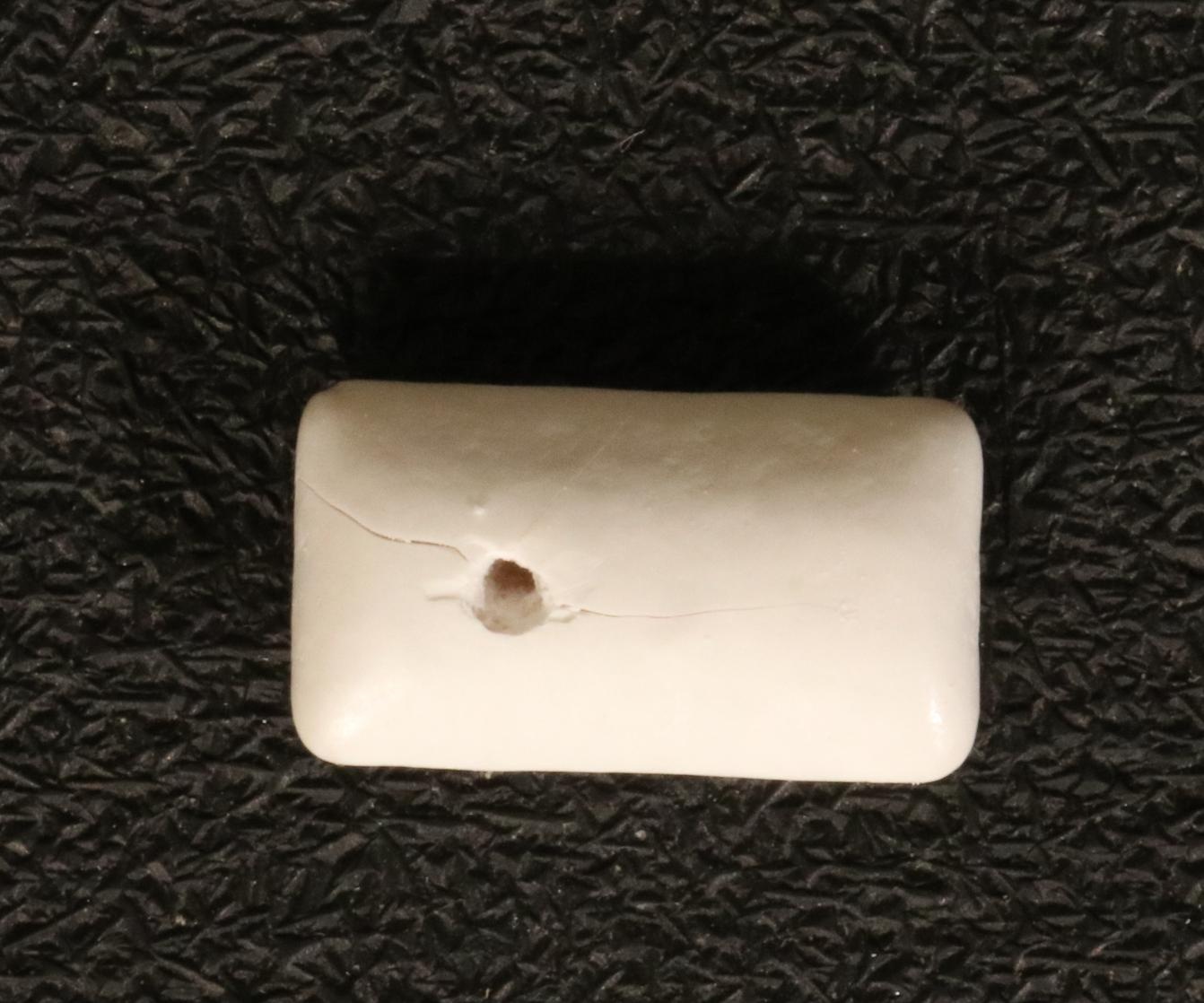}} & \multirow{3}{*}{\includegraphics[width=0.08\textwidth]{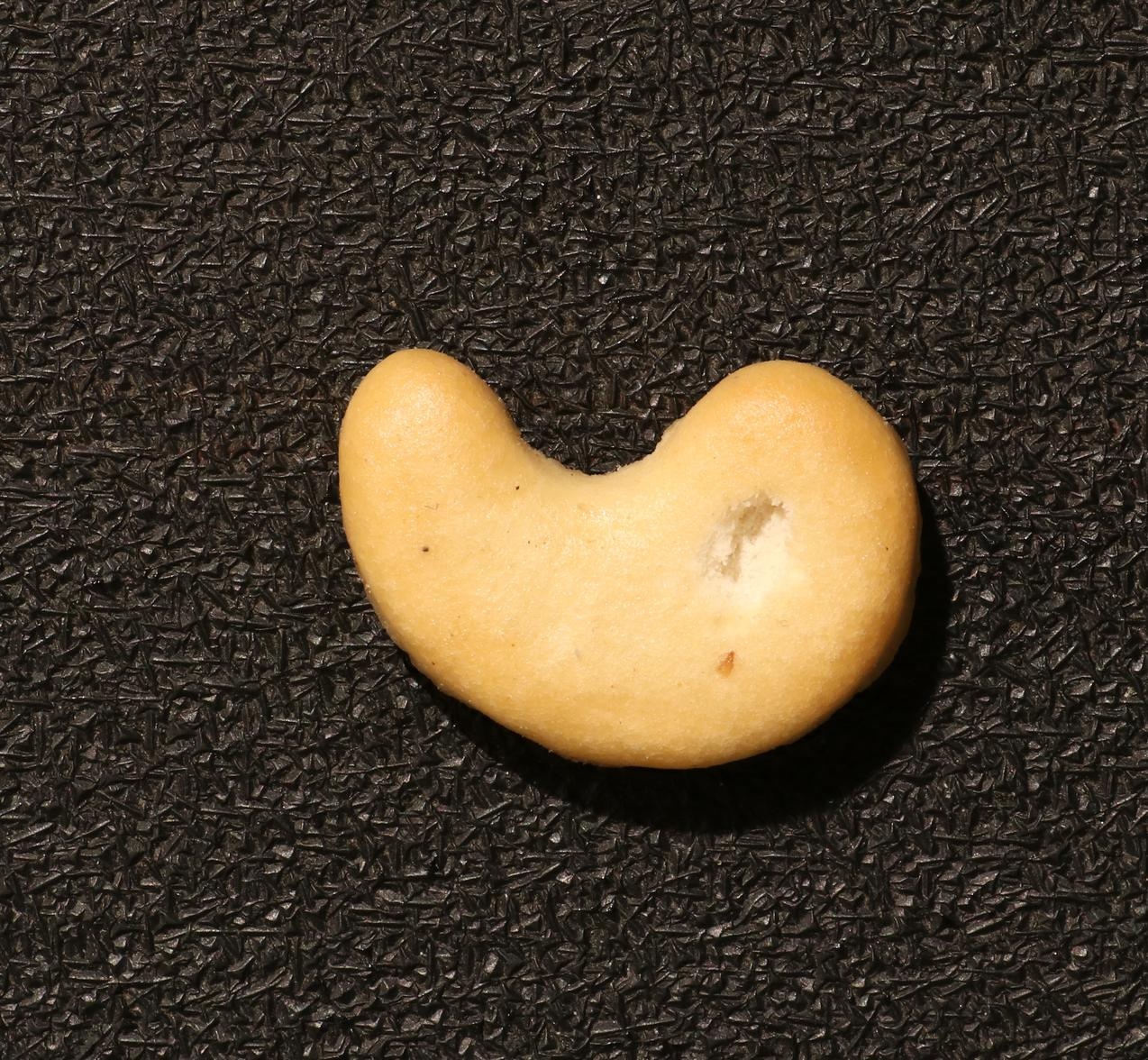}}  & unsupervised & 97.1(+0.0) & 98.8(+0.0) \\ 
    &&& supervised$_{\text{others}}$ & 98.6(\textcolor{red}{+1.5}) & 99.1(\textcolor{red}{+0.3}) \\ 
    &&& supervised & 98.7(\textcolor{red}{+1.6}) & 99.4(\textcolor{red}{+0.6}) \\
    \hline
    \multirow{3}{*}{pcb2} & \multirow{3}{*}{\includegraphics[width=0.10\textwidth]{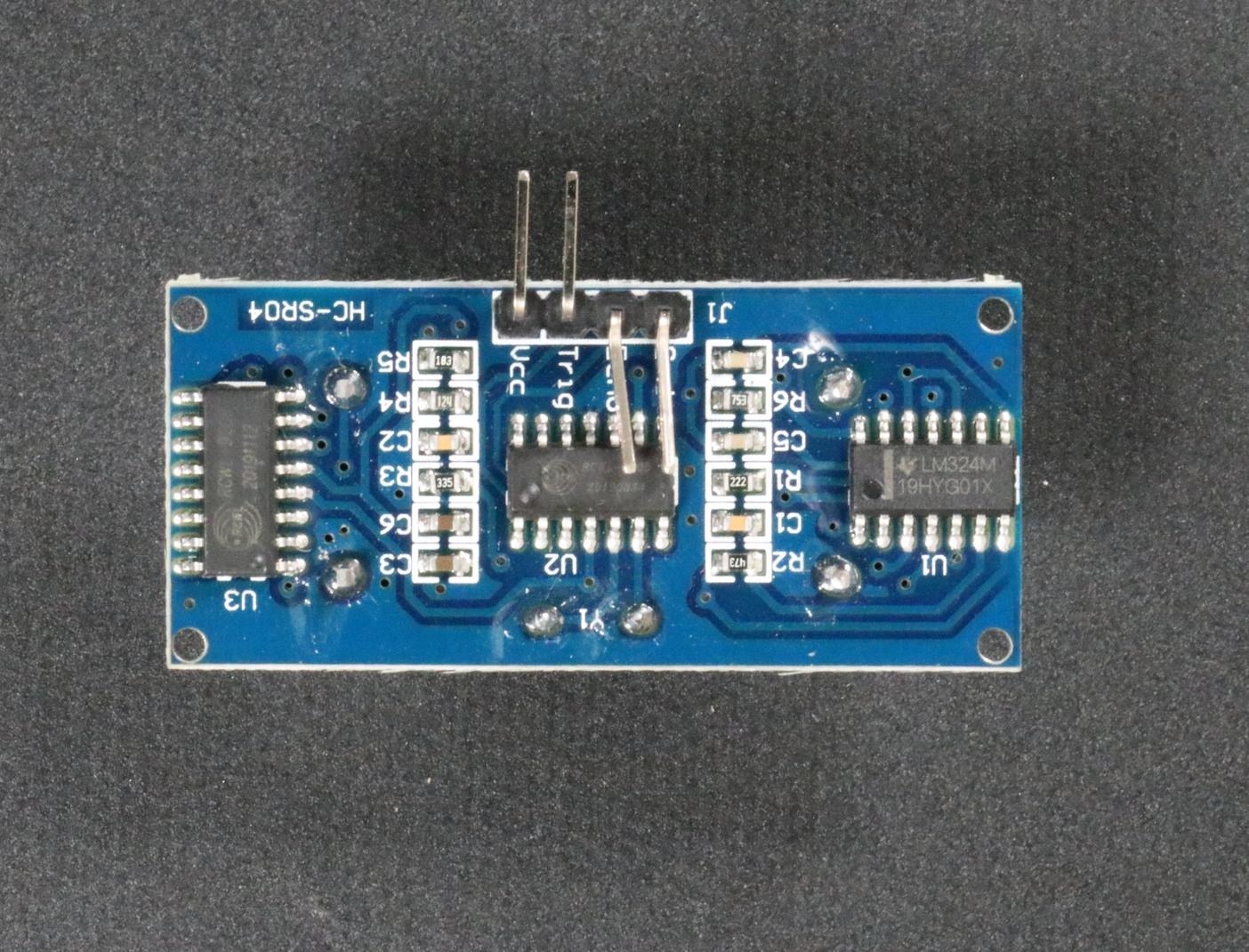}} & \multirow{3}{*}{\includegraphics[width=0.10\textwidth]{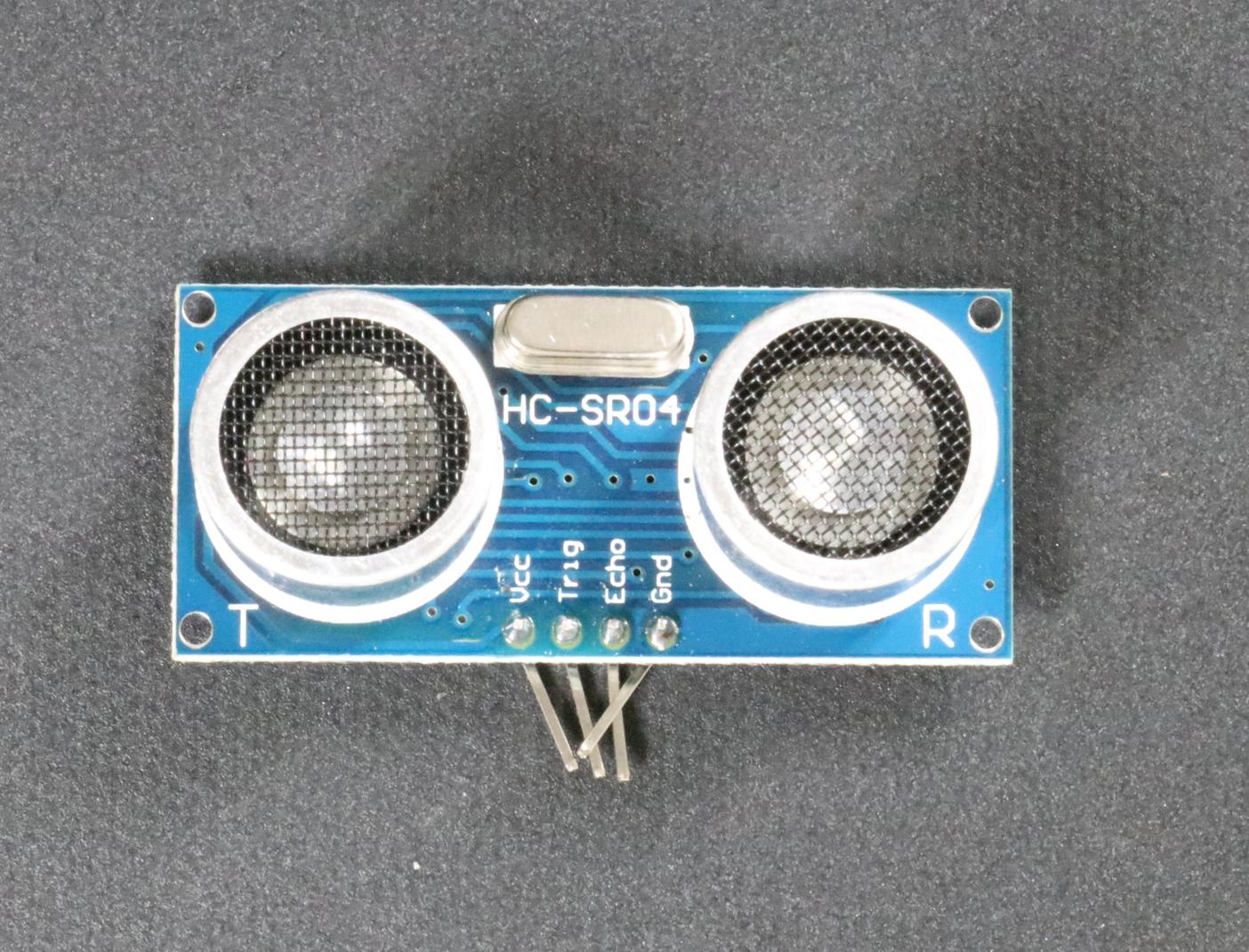}} & unsupervised & 94.1(+0.0) & 98.3(+0.0) \\ 
    &&& supervised$_{\text{others}}$ & 96.7(\textcolor{red}{+2.6}) & 98.5(\textcolor{red}{+0.2}) \\ 
    &&& supervised & 98.5(\textcolor{red}{+4.4}) & 99.4(\textcolor{red}{+1.1}) \\
    \hline
    \end{tabular}
\vspace{-0.5cm}
\end{table}

\begin{table}[ht]
    \centering
    \begin{minipage}[t]{0.46\linewidth}
        \centering
        {\includegraphics[width=\linewidth]{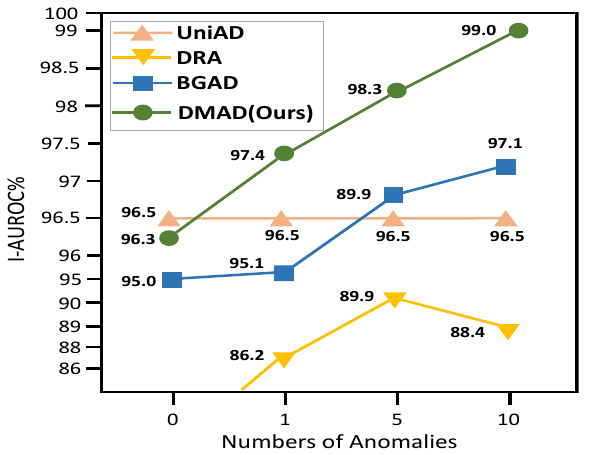}}
        \captionof{figure}{I-AUROC comparison for varying numbers of anomalies on MVTec.}
        \label{figure/linechart}
    \end{minipage}
    % \hspace{1pt}
    \begin{minipage}[t]{0.45\linewidth}
        \centering
        \raisebox{2pt}
        {\includegraphics[width=\linewidth]{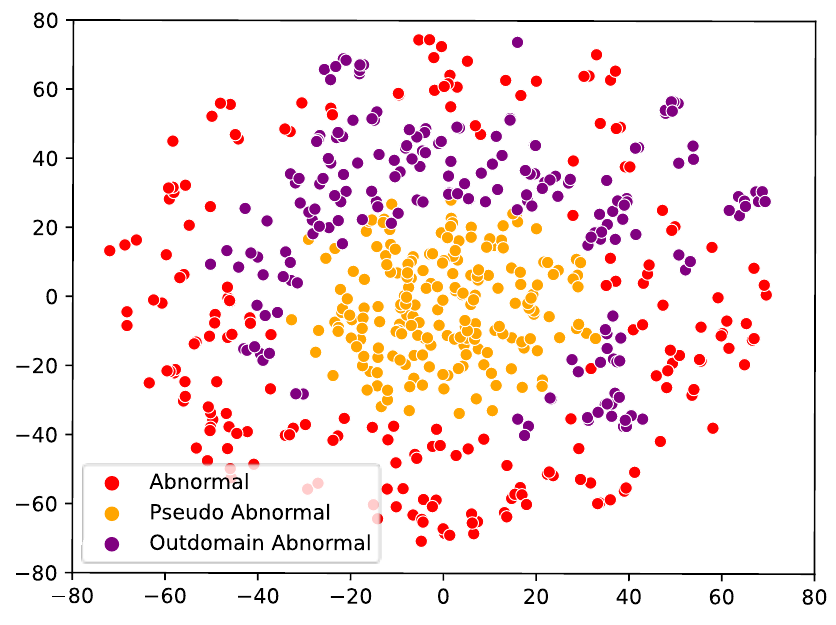}}
        \captionof{figure}{Visualization of three different abnormal features in memory bank.}
        \label{fig:scatter}
    \end{minipage}
    \vspace{-1cm}
\end{table}

\subsection{Analysis}
\textbf{Learning from other object's defects.}
We conducted a specific analysis on the VisA dataset, selected objects include "Macaroni1" (Multiple Instances), and "Chewinggum" and "Pcb2" (Single Instance), to investigate if the seen anomalies from other object could enhance the performance of the target object. The results are detailed in \Cref{tab:analysis}. Det. and Loc. represent the results of Image-level and Pixel-level AUROC, respectively. Any gains are highlighted in \textcolor{red}{red}. The setting include no anomalies used (unsupervised), anomalies used (supervised), and anomalies from other objects used (supervised$_\text{others}$). We consider the \textbf{unsupervised} setting as the baseline.
Initially, we introduced seen anomalies from all other classes, excluding the anomalies of the class itself for training, denoted as \textbf{supervised$_{\text{others}}$}, resulted in a performance boost for each class. This finding suggests that under the unified semi-supervised setting, similar defects across different objects can be beneficial, thereby validating our hypothesis and underscoring the importance of this setting. Furthermore, when we introduced the supervision information of the class itself (\textit{i.e.} \textbf{supervised}), we observed additional performance improvements.

\noindent\textbf{Real-world Anomaly Detection.}
A unified model is more practical for real-world anomaly detection, and in real-world scenes, the number of anomalies could be varied. DMAD could handle both unsupervised and semi-supervised scenarios and is benefits of growing anomalies, which is shown in \Cref{figure/linechart}.

\noindent\textbf{Intuitively visualization of Abnormal Memory Bank.}
DMAD use a three parts feature to comprised the abnormal memory bank, the outlier data augmented features, seen abnormal features, and the pseudo abnormal features generated by anomaly center sampling strategy. We use t-SNE to reduction the three anomaly features, which is shown in \Cref{fig:scatter}, we could see the new anomalies greatly enhances the diversity of abnormal.

\section{Conclusion}
In this study, we tackle real-world anomaly detection by introducing a novel framework, which we have named Dual Memory bank enhanced representation learning for Anomaly Detection (DMAD). DMAD is a unified framework capable of managing both unsupervised and semi-supervised scenarios within a multi-class setting. It employs a dual memory bank to compute the knowledge of normal and abnormal instances, which is then used to construct an enhanced representation for anomaly score learning. Evaluation results on the MVTec-AD and VisA datasets demonstrate the superior performance of DMAD.

\noindent\textbf{Limitation.} In practical industrial scenarios, the number of anomalies can vary significantly. This study merely simulates simplified scenes, setting the number of anomalies to 0 (unsupervised), 1, 5, and 10. Furthermore, per-pixel annotation for anomalies may not be available, with only the class label known. In such cases, a new approach needs to be employed to utilize these anomalies.

\noindent\textbf{Acknowledgements.}This work was supported by National Science and Technology Major Project (No. 2022ZD0118202), the National Science Fund for Distinguished Young Scholars (No.62025603), the National Natural Science Foundation of China (No. U21B2037, No. U22B2051, No. 62176222, No. 62176223, No. 62176226, No. 62072386, No. 62072387, No. 62072389, No. 62002305 and No. 62272401), and the Natural Science Foundation of Fujian Province of China (No.2021J01002,  No.2022J06001).

% \clearpage\mbox{}Page \thepage\ of the manuscript.
% \clearpage\mbox{}Page \thepage\ of the manuscript.
% \clearpage\mbox{}Page \thepage\ of the manuscript.
% \clearpage\mbox{}Page \thepage\ of the manuscript.
% \clearpage\mbox{}Page \thepage\ of the manuscript. This is the last page.
% Now we have reached the maximum length of an ECCV \ECCVyear{} submission (excluding references).
% References should start immediately after the main text, but can continue past p.\ 14 if needed.
% \clearpage  % TODO REVIEW/FINAL: This \clearpage needs to be removed from both review and camera-ready versions.

\bibliographystyle{splncs04}
\bibliography{main}
\end{document}